\documentclass[lettersize,journal,romanappendices]{IEEEtran}
\usepackage{amsmath,amsfonts}
\usepackage{algorithmic}
\usepackage{algorithm}
\usepackage{array}
\usepackage[caption=false,font=normalsize,labelfont=sf,textfont=sf]{subfig}
\usepackage{textcomp}
\usepackage{stfloats}
\usepackage{url}
\usepackage{verbatim}
\usepackage{graphicx}
\usepackage{cite}

\usepackage{amssymb}
\usepackage{amsthm}
\usepackage{multirow}
\usepackage{subfig}
\usepackage{makecell}
\usepackage{wrapfig}
\usepackage{xcolor}
\usepackage{colortbl}

\newcommand{\tabincell}[2]{\begin{tabular}{@{}#1@{}}#2\end{tabular}}
\newenvironment{sequation}{\small\begin{equation}}{\end{equation}}

\theoremstyle{plain}
\newtheorem{theorem}{Theorem}
\newtheorem{proposition}[theorem]{Proposition}

\theoremstyle{definition}

\theoremstyle{remark}

\begin{document}

\title{Relational Experience Replay: Continual Learning \\by Adaptively Tuning Task-wise Relationship}


\author{Quanziang Wang, Renzhen Wang, Yuexiang Li, Dong Wei, Hong Wang,\\ Kai Ma, Yefeng Zheng,~\IEEEmembership{Fellow,~IEEE}, Deyu Meng,~\IEEEmembership{Member,~IEEE}
\thanks{Quanziang Wang, Renzhen Wang, and Deyu Meng are with the School of Mathematics and Statistics, Xi'an Jiaotong University, Xi'an 710049, P.R. China.
}
\thanks{Yuexiang Li, Dong Wei, Hong Wang, Kai Ma, and Yefeng Zheng are with Tencent Jarvis Lab, Shenzhen 518052, P.R. China.}}

\markboth{Journal of \LaTeX\ Class Files,~Vol.~14, No.~8, August~2021}%
{Shell \MakeLowercase{\textit{et al.}}: A Sample Article Using IEEEtran.cls for IEEE Journals}


\maketitle

\begin{abstract}
Continual learning is a promising machine learning paradigm to learn new tasks while retaining previously learned knowledge over streaming training data. Till now, \textit{rehearsal-based} methods, keeping a small part of data from old tasks as a memory buffer, have shown good performance in mitigating catastrophic forgetting for previously learned knowledge. However, most of these methods typically treat each new task equally, which may not adequately consider the relationship or similarity between old and new tasks. Furthermore, these methods commonly neglect sample importance in the continual training process and result in sub-optimal performance on certain tasks. To address this challenging problem, we propose Relational Experience Replay (RER), a bi-level learning framework, to adaptively tune task-wise relationships and sample importance within each task to achieve a better `stability' and `plasticity' trade-off. As such, the proposed method is capable of accumulating new knowledge while consolidating previously learned old knowledge during continual learning. Extensive experiments conducted on three publicly available datasets (\textit{i.e.}, CIFAR-10, CIFAR-100, and Tiny ImageNet) show that the proposed method can consistently improve the performance of all baselines and surpass current state-of-the-art methods.
\end{abstract}


\section{Introduction}
\IEEEPARstart{D}{eep} neural networks trained offline have achieved excellent results in many computer vision tasks, such as classification \cite{VGG, resnet, DenseNet}, semantic segmentation \cite{PSPnet, MaskRCNN, PLOP}, and object detection \cite{FasterRCNN, YOLOv3, YOLOv4}. However, these models commonly struggle in continual learning (CL) scenarios where the knowledge is incrementally aggregated from data that are generated by a non-stationary distribution. Actually, the inability of the models is caused by the fact that they are inclined to overwrite the previously learned knowledge whenever new tasks come in. This is known as catastrophic forgetting \cite{catastrophic_forgetting} and is a key challenge in continual learning. 

\begin{figure}[t]
\begin{center}
\centerline{\includegraphics[width=\columnwidth]{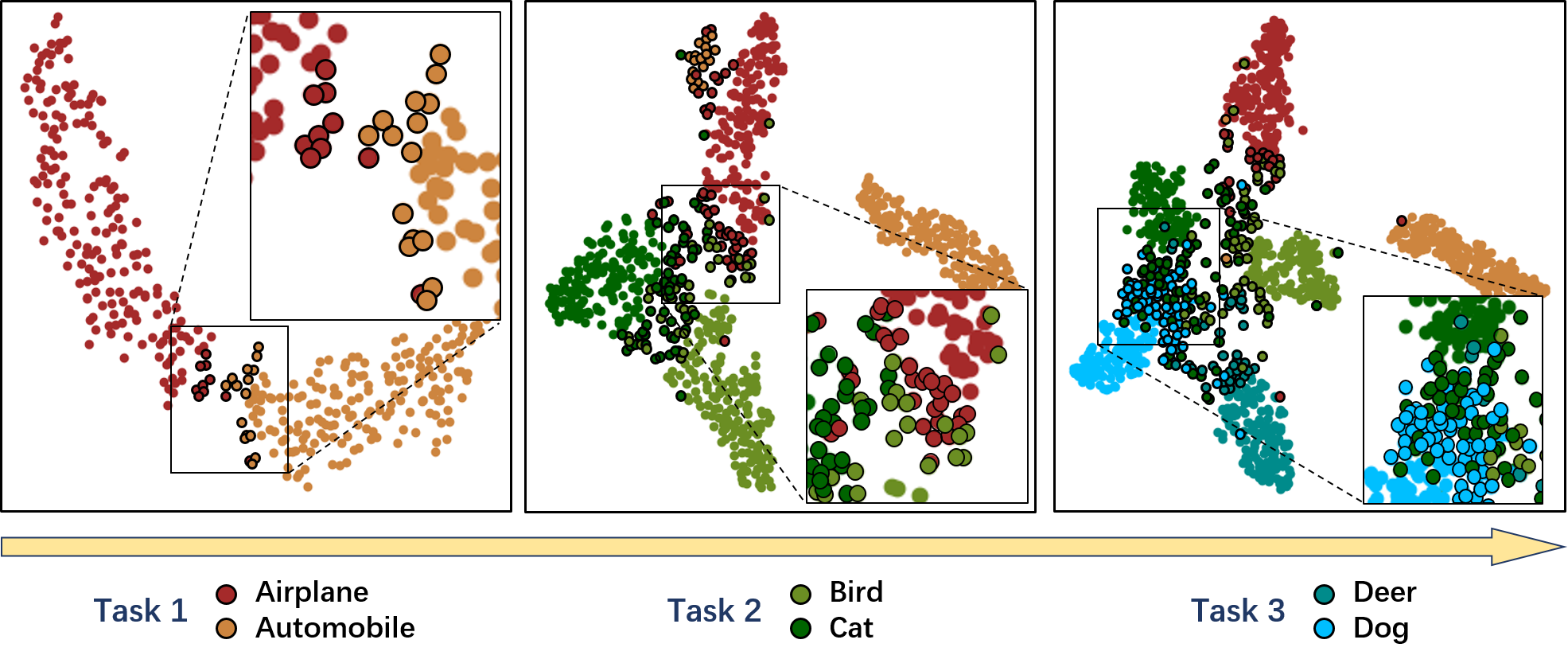}}
\caption{Illustration of two main factors affecting stability and plasticity in continual learning: (i) The relationship between the new and old tasks varies dynamically. For example, the class `bird' of task 2 is semantic-related to the `airplane' of task 1 due to the closer distance between these two clusters in the latent space. (ii) The importance of data points within each class is different for model training. The samples with black borders, located around the classification boundary, are more difficult/beneficial for the classification than other samples.}
\label{fig:joint}
\end{center}
\vspace{-3mm}
\end{figure}

Recently, various CL methods have been proposed to alleviate the catastrophic forgetting problem. Existing algorithms for CL can be roughly divided into three categories \cite{survey}:
1) \textit{Model-based} approaches \cite{PNN, PathNet, HAT, ConditionalChannelGated, DynamicallyExpandable}, which dynamically modify the architecture of the model to process the new task and use specific model structures for different tasks in the testing phase; 2) \textit{Regularization-based} approaches \cite{EWC, LwF, SI, oEWC, NeuronCalibration}, which propose constraints to maintain the model parameters that are relatively important for previously learned tasks; 3) \textit{Rehearsal-based} approaches \cite{ER1, ER2, icarl, GEM, DER, HAL, rainbow, ER-ACE, SSIL}, which store a small subset of examples from previously learned tasks in a memory buffer and replay them during the learning of new tasks.
In this study, we mainly focus on exploring the potential of \textit{rehearsal-based} methods, due to their relatively stable and effective performance in mitigating catastrophic forgetting.

Due to the lack of data from old tasks, except for a small memory buffer, \textit{rehearsal-based} methods require striking an appropriate balance between learning new information and retaining previous knowledge, which is referred to as the `stability-plasticity' dilemma \cite{stability_plasticity}.
Many \textit{rehearsal-based} methods, such as Experience Replay (ER) \cite{ER1, ER2} and DER++ \cite{DER}, primarily concentrate on leveraging stored information from previous tasks to mitigate model forgetting during the training of new tasks. However, these approaches have paid limited attention to modeling the intricate relationships between tasks.
For example, the loss function of ER is the sum of the losses of incoming new task samples and old task ones stored in the memory buffer: $\mathcal{L}_{new} + \lambda \mathcal{L}_{old}$, where $\lambda$ is a hyperparameter and it is often tuned manually by grid search at the beginning of the training process to balance the trade-off between stability and plasticity. 
This trade-off hyperparameter is generally required to be carefully pre-specified to guarantee a sound training tendency across the whole CL process.

Actually, as shown in Fig.~\ref{fig:joint}, under the settings of CL, the data streaming of new tasks inclines to induce the continual variation of data distribution. Therefore, the relationship between old and new tasks dynamically varies throughout the learning process. 
For example, the knowledge extracted from previous tasks involving cats may be disturbed by the new task of dog identification, which may lead to serious forgetting of cat classification and suboptimal performance for the learned model. Conversely, certain semantically unrelated classes (\emph{e.g.,} automobile) have little impact on cats in old tasks.
Obviously, the continual model should adapt its focus on old tasks (\emph{i.e.}, `stability') or new tasks (\emph{i.e.}, `plasticity') in real-time according to their dynamically varying relationship. 
However, it is difficult for existing methods to model such a complicated task-wise relationship by simply tuning and fixing hyperparameters at the beginning of network training. 
Moreover, previous studies do not extensively consider another intrinsic factor, \emph{i.e.,} sample importance within the same class on feature representation learning.
Concretely, easy-to-learn samples contain more representative knowledge for each class, while the `hard' ones, located around decision boundaries for classification, can contribute to the learning of the classifier \cite{Zhou_2020_CVPR} (refer to Fig.~\ref{fig:joint}).
From this perspective, it is essential to model the relationship between new and old tasks and sample importance within the same class.

To this end, we propose \textit{Relational Experience Replay} (RER), a bi-level learning framework to couple the aforementioned complicated task relationship and sample importance.
Concretely, the inner-loop optimization problems aim to address the `plasticity' by leveraging the examples from the newly coming task, while the outer-loop optimization problem of RER aims to address the issue of `stability' by minimizing the empirical risk of a batch of balanced data (including samples from old tasks stored in the memory buffer and ones from new tasks).
To take full advantage of the training samples by considering the inter-task relationship and sample importance, we design an explicit weighting function parameterized by a light-weight neural network (dubbed Relation Replay Net, or RRN), which maps pairwise abstract information of samples from new and old tasks to their corresponding loss weights. 
We theoretically prove that the RRN is updated based on the similarity of the average gradient between classes in the outer-loop optimization, indicating that our proposed RER can implicitly model task-wise relationships.
We conduct extensive experiments on different benchmark datasets, and the results verify the effectiveness of the proposed method in consistently improving various baselines for \textit{rehearsal-based} continual learning methods.
In summary, our contributions are mainly four-fold:

1) The proposed method takes two factors, \emph{i.e.}, task relationship across the whole continual learning process and sample importance within each class, into account for sample weight assignment. Such a design facilitates the model to deal with the `stability-plasticity' dilemma that plagues the continual learning paradigm. 

2) We theoretically prove that the proposed method can implicitly model the task relationship. Specifically, the updating formulation of the Relation Replay Net depends on the similarity between the gradient of each training sample and the averaged gradient of each class stored in the memory buffer.

3) As far as we know, in the continual learning problem, we are the first to propose an automated sample weighting strategy to adaptively assign a reasonable weight to each sample from the new and old tasks, which is more flexible than existing approaches based on manual tuning.

4) The proposed method can be applied to various \textit{rehearsal-based} continual learning baselines and consistently improves their performance under various settings.

This paper is organized as follows. Section~\ref{sec:related_work} provides a review of some related works and Section~\ref{sec:preliminaries} briefly introduces the setting and necessary notations for continual learning. Section~\ref{sec:relational_experience_replay} presents the proposed RER method in detail. Section~\ref{sec:experiments} then provides the experiments and analysis of our method. The paper is finally concluded in Section~\ref{sec:conclusion}.

\section{Related Work}
\label{sec:related_work}

\subsection{Continual Learning}

\noindent\textbf{Rehearsal-based Methods.}
The primary mechanism underlying \textit{rehearsal-based} methods \cite{ER1, ER2, icarl, GEM, DER, HAL, rainbow, ER-ACE, LUCIR} is using the information of data from old tasks to prevent forgetting while training new tasks.
Specifically, these methods typically involve saving a portion of data samples from old tasks as a memory buffer, which is subsequently used alongside new incoming data samples to train the model.
For example, GEM~\cite{GEM} formulates a quadratic programming problem to enforce orthogonality between the optimization direction for a new task and the previously stored in the memory buffer during training.
A-GEM \cite{A-GEM} relaxes the constraints in GEM by only restricting the dot product of the new and old sample gradients to be non-negative, so as to improve the computational efficiency of the algorithm.
iCaRL \cite{icarl} stores the most representative samples, \textit{i.e.}, located around the class center in the latent space as the memory buffer, and uses the Nearest Class Mean (NCM) classifier to mitigate the impact of feature representation changes. 
Rainbow Memory \cite{rainbow} constructs the memory buffer by sampling more representative samples, which is determined by the prediction confidence of the data after various augmentations.
In addition to the ground truth labels, DER++~\cite{DER} also saves data probabilities yielded by the model from previous epochs in the memory buffer as soft labels, which are used for distillation to further prevent forgetting.
On the flip side, \textit{rehearsal-based} methods commonly lead to class imbalance problems, as they rely on a small memory buffer to store a limited number of examples from old tasks. To address this problem, LUCIR~\cite{LUCIR} normalizes the predicted logits and imposes constraints on the features to correct the imbalance problem between new and old samples. ER-ACE~\cite{ER-ACE} calculates the loss function of the new and old tasks independently to alleviate the mutual interference between them.

\noindent\textbf{Other Continual Learning Methods.}
\textit{Model-based} approaches~\cite{PNN, PathNet, packnet, HAT, ConditionalChannelGated, DynamicallyExpandable} aim to endow the model with the ability to adapt to new tasks by automatically modifying its architecture. Although these methods demonstrate impressive performance in a variety of simulation experiments, their training, and optimization are known to be challenging and computationally demanding.
For example, PNN~\cite{PNN} saves all networks of previous tasks to avoid forgetting, which often occupies a large memory buffer and needs to train another network for the new task.
\textit{Regularization-based} approaches \cite{EWC, LwF, SI, VCL, oEWC, NeuronCalibration} design different constraints to prevent changing important parameters of previous tasks during training new tasks.
Typically, EWC~\cite{EWC} limits the updating of important parameters, which are selected by the Fisher matrix from a Bayesian perspective.
SI~\cite{SI} determines important parameters by the effect of the parameter change on the loss. LwF~\cite{LwF} saves previous model predictions of new task samples as soft labels to distill the extracted knowledge.
Despite their simplicity and effectiveness, these approaches also face challenges such as hyper-parameter tuning and sensitivity to the choice of regularization parameters.
Besides, continual learning also be adopted on many realistic problems, such as semantic segmentation~\cite{segment_1,segment_2,segment_tmm}, few-shot learning~\cite{few_shot1,few_shot2,few_shot_tmm}, and emotion detection~\cite{emotion_tmm1,emotion_tmm2}, etc. For more detailed information, we recommend referring to \cite{survey, survey2}.

\subsection{Sample Weighting Strategy}
In terms of sample weighting, our method is closely related to L2RW~\cite{L2RW} and Meta Weight Net (MW-Net)~\cite{meta_weight_net}. These approaches involve training a classification network on a noisy label dataset and being optimized through a meta-learning strategy to mitigate the impact of noisy labels. This meta-learning process is guided by a small clean dataset, referred to as the meta set.
However, the challenges of CL are different from the noisy label problem, and especially the unstable data flow and severe catastrophic forgetting make it difficult to directly apply these sample weighting methods to CL.
To mitigate this problem, we propose a novel pair-wise sample weighting strategy to model task relationships and it does not require any additional high-quality data as the meta set like L2RW or MW-Net due to the limitation of continual learning settings.
To our knowledge, this should be the first work to use the automatic sample weighting strategy for the continual learning problem.

\section{Preliminaries}
\label{sec:preliminaries}

In this section, we briefly introduce the settings of the continual learning problem and two main \textit{rehearsal-based} baselines.
In continual learning, the model $f$ is required to learn a stream of tasks $\mathcal{T}=\{\mathcal{D}_1, \mathcal{D}_2, \cdots\}$ and for each task $\mathcal{D}_t$, only a few samples can be stored in a memory buffer $\mathcal{M}$, where the buffer size is denoted as $M$.
Typically, the model $f: \mathcal{X} \to \mathbb{R}^{C_t}$ is a classification neural network parameterized by $\theta$, where $C_t$ denotes the total number of classes across tasks 1 to $t$. 
In this study, we mainly focus on the \textit{Class Incremental} (Class-IL) and \textit{Task Incremental} (Task-IL) settings of continual learning.
Specifically, in Task-IL, the task boundary is clear, \emph{i.e.,} 
the task ID $t$ is known during training and testing.
Therefore, the model can feed the input data to the corresponding task-specific classifier using masks \cite{DER} based on its task ID.
Contrastively, the task ID is unreachable in Class-IL, which is more difficult.

\subsection{Experience Replay (ER)} 
ER is a fundamental \textit{rehearsal-based} approach that aims to mitigate the forgetting of previously learned tasks while learning new ones by replying to samples stored in the memory buffer. Specifically, during each task training, it samples a batch of data $\mathcal{B}^D = \{(x^D_i, y^D_i)\}^{B}_{i=1}$ from the current $t$-th task $\mathcal{D}_t$ and another batch of data $\mathcal{B}^M = \{(x^M_i, y^M_i)\}^{B}_{i=1}$ from the memory buffer $\mathcal{M}$ with the same batch size $B$, where $x_i$ and $y_i$ represent an example and its corresponding label, respectively. The loss function of the model can be formulated as:
\begin{equation}
    \mathcal{L}^{tr}(\theta) = 
    \frac{1}{B} \sum^{B}_{i=1} 
    \lambda^D {L}_{CE}(x^D_i; \theta) + \lambda^M {L}_{CE}(x^M_i; \theta),
\label{eq:ER}
\end{equation}
where the loss function ${L}_{CE}$ is cross-entropy (CE) loss.

In Eq.~(\ref{eq:ER}), the hyperparameters $\lambda^D$ and $\lambda^M$ represent the weights assigned to the loss functions for the new task and the memory buffer, respectively. Typically, a common setting is $\lambda^D=1$ and $\lambda^M=\lambda$. The value of $\lambda$ is a crucial hyperparameter that must be preset manually before training. A larger $\lambda$ value can enhance the model's plasticity to improve performance on the new task, while a smaller $\lambda$ value can emphasize stability and prevent forgetting the previous tasks. However, setting a fixed weight for $\lambda$ is not always optimal for all datasets, and tuning it for each incoming task poses a significant challenge.

\subsection{Dark Experience Replay++ (DER++)}
DER++ [26], an extension of ER, adds an additional knowledge distillation loss $\mathcal{L}_{KD}$ between the prediction probabilities of the current model and those of the previous models for samples in the memory buffer to further alleviate the forgetting compared with ER. Formally, the training objective is
\begin{sequation}
    \mathcal{L}^{tr}(\theta) \!\!= \!\!
    \frac{1}{B} \!\sum_{i=1}^{B}\! 
     \lambda^D {L}_{CE}(x^D_i; \theta)
     + \lambda^{M} {L}_{CE}(x^M_i; \theta)
     + \gamma^{M} {L}_{KD}(x^M_i; \theta),
\label{eq:DER}
\end{sequation}
where the weight for the new task is consistent with ER ($\lambda^D = 1$), and the weights for the loss function of memory buffer $[\lambda^{M}, \gamma^{M}]$ are two hyperparameters that are typically manually set by grid search.
Note that these hyperparameters are interrelated and play a crucial role in determining the `stability-plasticity' trade-off of the model. However, similar to ER, the two parameters are also required to be preset and fixed in the whole continual learning process. This poses a challenge on how to automatically balance `stability' and `plasticity' during training across different continual learning scenarios.

\section{Relational Experience Replay}
\label{sec:relational_experience_replay}

\begin{figure*}[!t]
\centering
\includegraphics[width=0.9\textwidth]{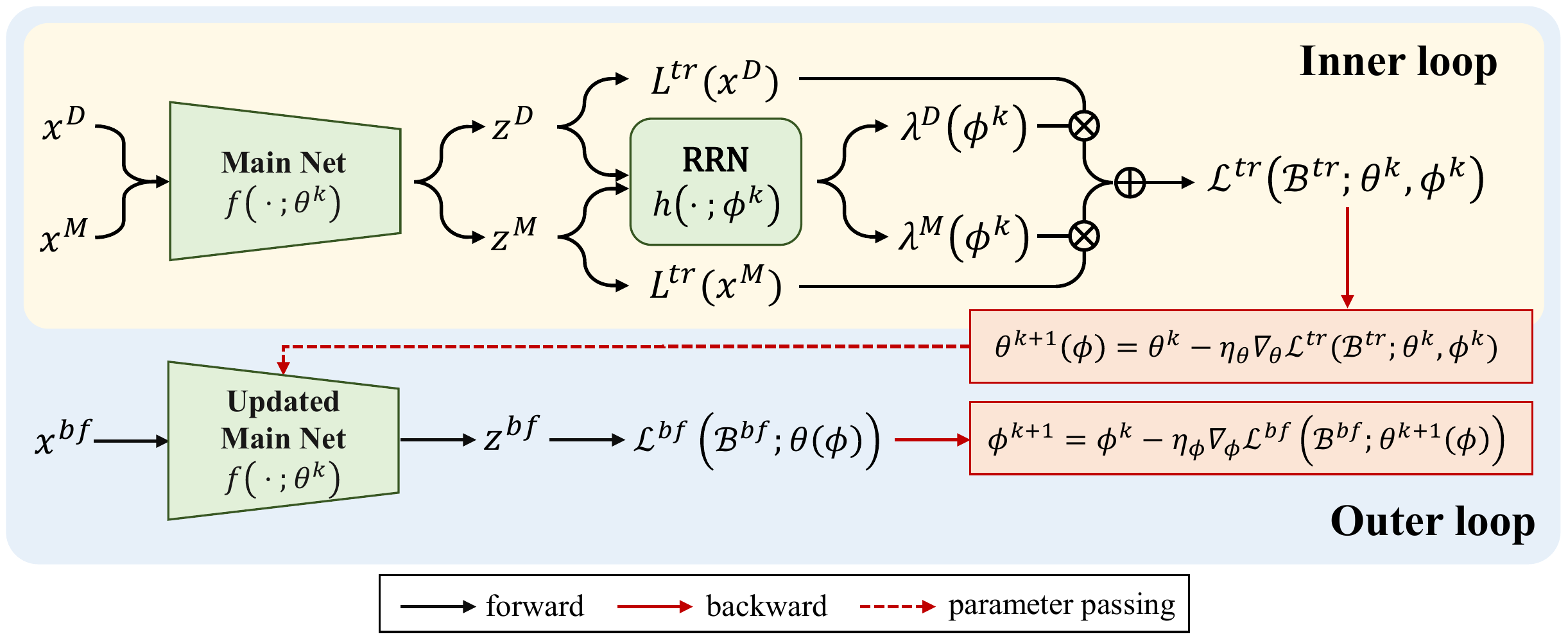}
\caption{The main illustration of the proposed Relational Experience Replay at $k$-th iteration. In the inner loop, the Main Net is trained on the batch of paired data $\mathcal{B}^{tr}$.
In the inner loop, the Relation Replay Net (RRN) generates the weights of the paired batch of training samples $\mathcal{B}^{tr}$, and the Main Net is updated by SGD based on the weighted training loss.
In the outer loop, given the updated parameters of the Main Net, RRN is trained with the batch of data sampled from the memory buffer, which is a relatively balanced dataset.}
\label{fig:network-update}
\end{figure*}

Most existing works alleviate the model forgetting problem by storing more information about old tasks, which can potentially result in an increased demand for data storage. In this study, we aim to enhance continual learning from a different perspective---assigning proper weights to training samples based on the following factors: 1) task relationship between new and old tasks, and 2) sample importance within each task for network training.
To account for the first factor, we consider that a sample from the memory buffer containing similar or semantically-relevant information to the new task samples is more prone to suffer from disturbance, potentially leading to more forgetting.
In this situation, the model should probably pay more attention to this memory buffer sample (\emph{i.e.,} assigning larger sample weights to this old sample or decreasing weights of new task samples) to alleviate the forgetting issue.
Vice versa, the model should properly assign larger weights to the new task samples to better learn new knowledge without disturbing old tasks too much.
As for the second factor, the data points from the same class also tend to make different contributions to model training. The easy-to-learn samples deliver more representative feature information of their classes, while the `hard' samples are more conducive to refining the classification boundaries.

As mentioned above, the loss weights $\Lambda$, which control the balance between `stability' and `plasticity' during training across different continual learning scenarios, should be dynamically assigned to different samples instead of being manually tuned and fixed~\cite{DER}.
To this end, we propose a Relation Replay Net (RRN) that extracts the interaction knowledge of the new and old samples and generates their corresponding weights $\Lambda$ dynamically. It facilitates the main classification network (dubbed Main Net) to achieve a better trade-off between `stability' and `plasticity'. Intuitively, the RRN depends on the training state of the Main Net, and the sample weights generated by the RRN in turn affect the training of the Main Net. Therefore, instead of the naive end-to-end training, we adopt a bi-level learning framework to jointly optimize the RRN and the Main Net, and the superiority of the bi-level optimization will be demonstrated in Sec.~\ref{subsec:discussion}.
For simplicity, we initially apply the proposed approach to ER (termed Relational Experience Replay, or RER), and the implementation based on more baselines can be found in Appendix~\ref{appendix:apply-other-baselines}.

\subsection{Overview} 
\label{sec:overview}
The RER framework, as depicted in Fig.~\ref{fig:network-update}, consists of two key components: Main Net $f(\cdot; \theta)$ responsible for continual learning, which can be any commonly used backbone architecture, and RRN $h(\cdot; \phi)$ utilized for estimating sample weights.

On the one hand, a CL model must be able to adapt to different new tasks based on the knowledge learned from the old tasks, \emph{i.e.}, task relationship would affect the `plasticity' as aforementioned. Furthermore, each new task sample carries a different degree of influence on the model's plasticity, highlighting the importance of sample weights in fine-tuning the model's attention.
To model this dynamic `plasticity', we assign weights for pair-wise training samples consisting of new and old task samples to train the Main Net, where the weights are generated by the RRN based on their relationships.
Concretely, we combine the new task batch $\mathcal{B}^D$ and the memory batch $\mathcal{B}^M$ to pair-wisely construct training sample batch $\mathcal{B}^{tr}=\{(x^D_i, x^M_i)\}^{B}_{i=1}$, where we omit the labels for notation convenience.
Then, the weighted loss function for the Main Net $f$ formulates the inner loop optimization problem of the bi-level learning framework, that is:
\begin{equation}
\begin{split}
\small
    \theta^*(\phi) &= \mathop{\arg\min}_\theta \mathcal{L}^{tr}(\mathcal{B}^{tr};\theta,\phi) \\
    &\triangleq \frac{1}{B} \sum^{B}_{i=1} 
    \lambda^D_i(\phi) L^{tr}(x^D_i;\theta) + \lambda^M_i(\phi) L^{tr}(x^M_i;\theta),
\small
\end{split}
\label{eq:inner-loss}
\end{equation}
where $L^{tr}$ is CE loss based on ER, and the sample weights $\Lambda=\{\lambda^D_i, \lambda^M_i\}^{B}_{i=1}$ of these data pairs are generated by the RRN automatically.

On the other hand, the model should possibly alleviate the forgetting issue while learning new tasks. The proposed RRN should pay more attention to `stability' to prevent the Main Net from focusing too much on new tasks.
We thus formulate the outer loop optimization problem over memory buffer data, making the Main Net returned by optimizing the inner loop one in Eq.~(\ref{eq:inner-loss}) acts as a stable consolidation of knowledge from the learned tasks \footnote {Note that the parameter $\phi$ of the Relation Replay Net is regarded as a hyper-parameter of that of the Main Net.}, \textit{i.e.},
\begin{equation}
\begin{split}
\small
    \phi^* &= \mathop{\arg\min}_\phi \ \mathcal{L}^{bf}(\mathcal{B}^{bf}; \theta^*(\phi))\\
    &\triangleq \frac{1}{B} \sum^{B}_{i=1} L^{bf}(x_i;\theta^*(\phi)), 
\small
\label{eq:outer-loss}
\end{split}
\end{equation}
where the $L^{bf}(x; \theta)$ can be simply adopted as CE loss, and $\mathcal{B}^{bf}$ is another batch sampled from the memory buffer $\mathcal{M}$, which is different from the inner-loop training batch $\mathcal{B}^M$.
By dynamically generating specific weights for different samples, the RRN can help the Main Net to better balance the trade-off between new and old tasks, thereby facilitating the continual learning process.

Furthermore, the design of the RRN should consider the following two factors:
1) the RRN should take the \textbf{interaction information} between the new and old tasks into account;
2) the input of the RRN should be `\textbf{information abundant}' to ensure that the RRN could extract useful knowledge to generate meaningful weights.
With this aim, we propose to pair each new task sample with a corresponding sample from the memory buffer and use their respective abstract information as inputs to the RRN. Specifically, for each sample pair $x_i^D$ from the new task and $x_i^M$ from the memory buffer, we take their losses $L^{D,M} = \left[L^{tr}(x^D_i), L^{tr}(x^M_i)\right]$, and logit norm $||z^{D,M}|| = \left[\Arrowvert z^D_i \Arrowvert_2, \Arrowvert z^M_i \Arrowvert_2\right]$ as the inputs of the RRN, where $z=f(x;\theta)$ is the predicted logits. By taking into account both label and feature information, the RRN can capture the semantic and distributional similarities between the paired samples and generate appropriate weights for the Main Net to balance the importance of new and old tasks during training. Thus, the paired sample weights generated by the RRN can be written as:
\begin{equation}
    \left[\lambda^D_i(\phi), \lambda^M_i(\phi)\right]
    = h \left( L^{D, M}, \Arrowvert z^{D, M} \Arrowvert_2; \phi \right).
\label{eq:relation-net}
\end{equation}

In summary, the proposed RER forms a bi-level learning framework to simultaneously model the `plasticity' and `stability' in the inner and outer loop optimization problems, respectively.
The RRN, guided by a relatively balanced set $\mathcal{M}$, automatically generates sample weights to refine the optimization direction of the Main Net.
Therefore, it can facilitate the Main Net optimization, leading to a better trade-off between new and old tasks.
In the inference stage, we can directly predict the testing images by the Main Net without passing the RRN.
Note that our method can be easily adapted to other \textit{rehearsal-based} baselines, such as ER-ACE~\cite{ER-ACE} and DER++~\cite{DER} by some simple modification (please refer to Section~\ref{subsec:exp-results} and Appendix~\ref{appendix:apply-other-baselines}).

\subsection{Optimization Procedure}
\label{sec:update_meta_net}

Since it is difficult to find closed-form solutions, the optimization of $\theta$ and $\phi$ as shown in Eq.~(\ref{eq:inner-loss}) and Eq.~(\ref{eq:outer-loss}) depends on two nested loops, which is computationally expensive.
Considering computational efficiency and the large scale of data to be processed, we adopt an alternative online gradient-based optimization strategy to solve the proposed bi-level learning framework.

\medskip
\noindent\textbf{Updating $\theta$:}
Referring to Eq~(\ref{eq:inner-loss}), given the parameter $\phi^{k}$ of RRN at iteration step $k$, we optimize the parameter $\theta$ of Main Net by one-step gradient descent:
\begin{equation}
\small
    \theta^{k+1}(\phi) = 
    \theta^{k} - \eta_\theta \triangledown_\theta \mathcal{L}^{tr}(\mathcal{B}^{tr};\theta^k,\phi^k),
\label{eq:inner-update}
\end{equation}
where $\eta_\theta$ is the inner-loop learning rate. Note that the updated parameter $\theta^{k+1}(\phi)$ is actually a function of $\phi$.

\medskip
\noindent\textbf{Updating $\phi$:}
With Main Net parameter $\theta$, we can optimize RRN parameter $\phi$ by Eq.~(\ref{eq:outer-loss}) given $\theta^{k+1}(\phi)$ by the following formulation:
\begin{equation}
\begin{small}
    \phi^{k+1} 
    = \phi^{k} - \eta_\phi \triangledown_\phi \mathcal{L}^{bf}(\mathcal{B}^{bf}; \theta^{k+1}(\phi^k)),
\end{small}
\label{eq:outer-update}
\end{equation}
where $\eta_\phi$ is the outer-loop learning rate.
More details of the gradient calculation can be found in Appendix~\ref{appendix:update_meta_net}.

\begin{algorithm}[tb]
   \caption{Relational Experience Replay training algorithm}
   \label{alg:meta_training}
   
\textbf{Input}: new task data $\mathcal{D}_t$, memory buffer $\mathcal{M}$\\
\textbf{Output}: Main Net and Relation Replay Net (RRN) parameters $\{\theta, \phi\}$

\begin{algorithmic}[1] 
    \WHILE{$\mathcal{D}_t \neq \emptyset$} 
    \WHILE{$k<{Iter}_{max}$}
    \STATE Sample a new task batch $\mathcal{B}^{D} \in \mathcal{D}_t$ and a buffer batch $\mathcal{B}^{M} \in \mathcal{M}$
    \STATE Construct a paired training batch $\mathcal{B}^{tr} \gets \mathcal{B}^{D}$ and $\mathcal{B}^{M}$
    \STATE Calculate the inner-loop loss by Eq.~(\ref{eq:inner-loss})
    \STATE Update $\theta^{k}$ by Eq.~(\ref{eq:inner-update})
    \STATE Sample another buffer batch $\mathcal{B}^{bf} \in \mathcal{M}$
    \STATE Calculate the outer-loop loss by Eq.~(\ref{eq:outer-loss})
    \STATE Update $\phi^{k}$ by Eq.~(\ref{eq:outer-update})
    \STATE $k++$.
    \ENDWHILE

    \ENDWHILE
\end{algorithmic}
\end{algorithm}

   



\subsection{Theoretical Analysis}

According to Eqs.~(\ref{eq:inner-update}) and (\ref{eq:outer-update}), we have the following proposition to further reveal how the proposed method models the task-wise relationship.
\begin{proposition}
    Let $g^{bf}(x) = \left.\frac{\partial L^{bf}(x; \theta)}{\partial \theta}\right|_{\theta^k}$ and $g^{tr}(x) = \left.\frac{\partial L^{tr}(x; \theta)}{\partial \theta}\right|_{\theta^{k}}$ denote the gradients of the buffer sample and the training sample with respect to the parameter $\theta$, respectively.
    Then the updating formulation of $\phi$ presented in Eq.~(\ref{eq:outer-update}) can be reformulated as
    \begin{equation}
    \begin{small}
        \theta^{k+1}(\phi) = \theta^{k} + \frac{\eta_\theta \eta_\phi}{B} 
        \sum_{j=1}^{B} G(j) \cdot 
        \left. \frac{\partial h_j(\phi)}{\partial \phi} \right |_{\phi^{k}},
    \end{small}
    \label{eq:phi-update}
    \end{equation}
    where the gradient $\frac{\partial h_j(\phi)}{\partial \phi} = \left[\frac{\partial \lambda^D_j(\phi)}{\partial \phi}, \frac{\partial \lambda^M_j(\phi)}{\partial \phi}\right]^T$, and the coefficient $G(j)$ is
    \begin{equation}
    \begin{small}
        G(j) = \frac{1}{B}
        \sum_{c=1}^{C_t}
        \left( \sum_{i=1}^{B_c} g^{bf}(x_i)\right )
        \left[ g^{tr}(x^D_j), \quad g^{tr}(x^M_j) \right].
    \end{small}
    \label{eq:G}
    \end{equation}
    \label{proposition:class_similarity}
\end{proposition}

In Proposition~\ref{proposition:class_similarity}, we demonstrate that the task-wise relationship in the RRN can be implicitly modeled through the SGD updating process. This is achieved by computing the coefficient $G(j)$, which represents the mean dot product similarity between the gradient of buffer samples and that of training samples within each class. This measure effectively captures the relationship between the new task and the previously seen classes from a gradient perspective. The proof can be found in Appendix~\ref{appendix:update_meta_net}.

\subsection{Implementation Details}
\label{subsec:implementation}
 
\begin{figure}[!t]
\centering
\includegraphics[width=0.4\textwidth]{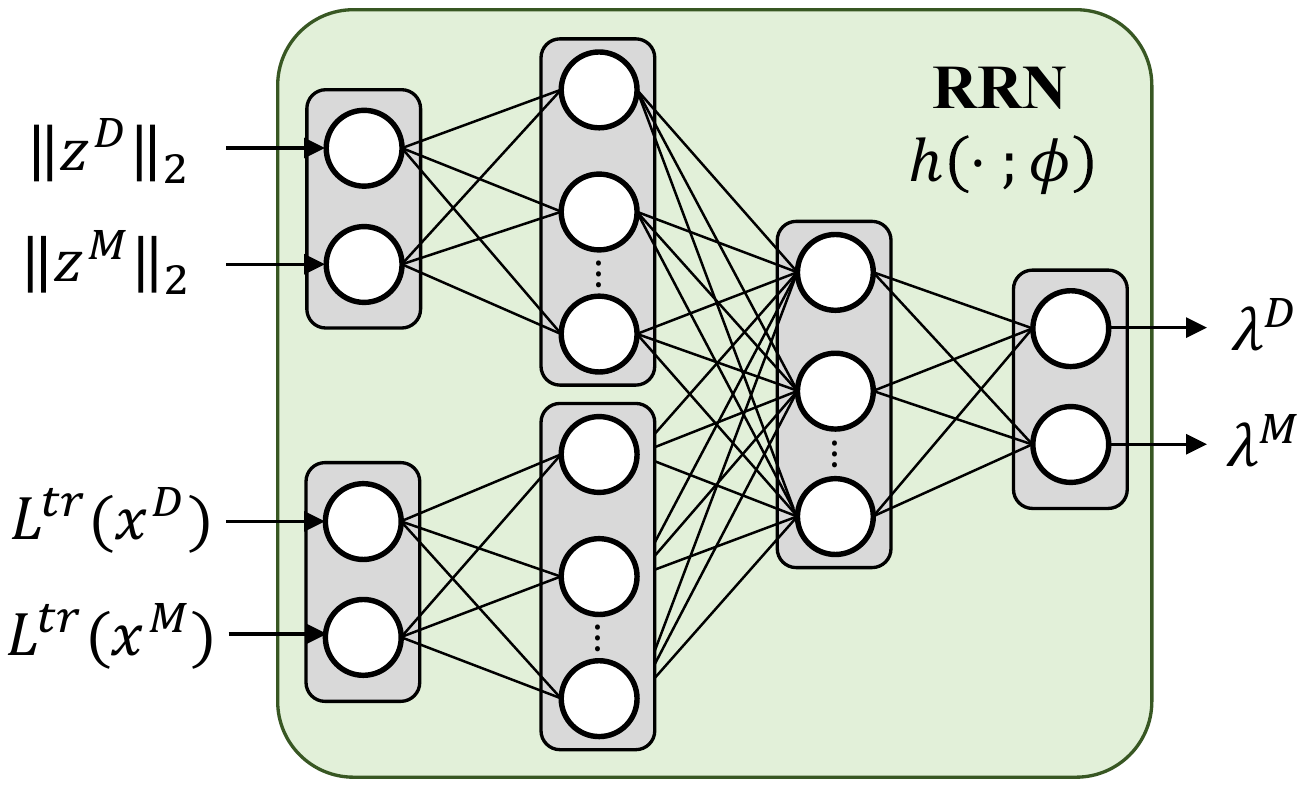}
\caption{The architecture of the Relation Replay Net (RRN).}
\label{fig:relation-net}
\vspace{-8mm}
\end{figure}

\noindent\textbf{The Architecture of the Relation Replay Net.}
We propose a two-hidden-layer neural network as the RRN. Intuitively, the first layer consists of two individual linear layers to extract the interaction information of loss values and the logit norms for paired training samples, respectively. The second layer is a linear layer that merges the extracted information to generate weights for the sample pairs automatically. Fig.~\ref{fig:relation-net} illustrates the architecture of RRN where we set the number of hidden units as 16 for computation efficiency.

\medskip
\noindent\textbf{Training Details.}
When training a new task, the RRN aims to generate reasonable sample weights for both new task samples and buffer samples to improve the trade-off between `stability' and `plasticity', which needs a warm-up stage to explore the relationship between the new and old tasks. Specifically, in the warm-up stage (stage length denoted as ${Iter}_{warm}$), we update the RRN by Eq.~(\ref{eq:outer-update}) and update the Main Net via preset fixed weights like the previous methods.
After the warm-up stage, we turn to use the sample weights generated by the RRN to guide the Main Net training rather than prefixed weights.
In addition, to reduce the calculation burden, we only update the RRN once while updating the Main Net for multiple steps (step number denoted as $Interval$). The specific choices of ${Iter}_{warm}$ and $Interval$ are presented in Section~\ref{subsec:ablation}.

\section {Experimental Results}
\label{sec:experiments}

To validate the effectiveness of our method, we conduct extensive experiments on three publicly available datasets, \emph{i.e.,} {\bf CIFAR-10}, {\bf CIFAR-100} \cite{cifar} and {\bf Tiny ImageNet} \cite{tinyimg}, and present thorough ablation analysis to gain insight into our method in this section.

\subsection{Experimental Setting}
For CIFAR-10, we divide the 10 classes into five tasks and each task is a binary classification problem. Both CIFAR-100 and Tiny ImageNet datasets are divided into 10 tasks, where each task is a 1-of-10 and 1-of-20 classification problem, respectively.
To ensure a robust and reliable evaluation, the framework was trained for 50 epochs on each task across all datasets following DER++~\cite{DER}.

As aforementioned, this study mainly focuses on the configurations of Class-IL and Task-IL. The detailed experimental settings are consistent with DER++ \cite{DER} for a fair comparison.
In our experiments, we evaluate the model by two common-used metrics, \textit{i.e.}, Average Accuracy (ACC) and Backward Transfer (BWT) ~\cite{GEM, DER}.
We repeat each experiment five times to reduce the randomness of network training.
Similar to \cite{DER}, the total size of the memory buffer remains constant throughout the entire training process.

The proposed method is implemented with the PyTorch platform \cite{pytorch}. The backbone of Main Net is the widely-used ResNet-18 \cite{resnet}.
In the inner loop, the Main Net is optimized by SGD with an initial learning rate of $0.03$ for all datasets, and in the outer loop, Adam \cite{adam} is adopted to optimize RRN, where the initial learning rate is set as 0.001 with a weight decay of $10^{-4}$.

\begin{table*}[t]
\caption{Comparison with three different baselines and other state-of-the-art methods on the CIFAR-10 and Tiny-ImageNet. The results of our method are shown in gray cells and the better results are presented in \textbf{bold}.}
\label{tab:baseline}
\centering
\resizebox{0.98\textwidth}{!}{
\begin{tabular}{cccccccccc}
\hline
\multirow{3}{*}{Buffer Size} & \multirow{3}{*}{Method} & \multicolumn{4}{c}{CIFAR-10}                                                                    & \multicolumn{4}{c}{Tiny-ImageNet}                                                               \\ \cline{3-10} 
                      &                         & \multicolumn{2}{c}{Class-IL}                   & \multicolumn{2}{c}{Task-IL}                    & \multicolumn{2}{c}{Class-IL}                   & \multicolumn{2}{c}{Task-IL}                    \\ \cline{3-10} 
                      &                         & ACC                   & BWT                    & ACC                   & BWT                    & ACC                   & BWT                    & ACC                   & BWT                    \\ \hline
- & Upper bound  & 92.20  \scriptsize ±0.15 & -                      & 98.31  \scriptsize ±0.12 & -                      & 59.99  \scriptsize ±0.19 & -                      & 82.04  \scriptsize ±0.10 & -                      \\
- & Lower bound    & 19.62  \scriptsize ±0.05    & -96.39  \scriptsize ±0.12   & 61.02  \scriptsize ±3.33    & -46.24  \scriptsize ±2.12   &  7.92  \scriptsize ±0.26    & -76.73  \scriptsize ±0.08   & 18.31  \scriptsize ±0.68   & -64.97  \scriptsize ±1.70 \\ \hline
- & oEWC   & 19.49  \scriptsize ±0.12 & -91.64  \scriptsize ±3.07 & 68.29  \scriptsize ±3.92 & -29.13  \scriptsize ±4.11 & 7.58  \scriptsize ±0.10 & -73.91  \scriptsize ±0.79 & 19.20  \scriptsize ±0.31 & -59.86  \scriptsize ±0.42          \\
- & SI     & 19.48  \scriptsize ±0.17 & -95.78  \scriptsize ±0.64 & 68.05  \scriptsize ±5.91 & -38.76  \scriptsize ±0.89 & 6.58  \scriptsize ±0.31 & -67.91  \scriptsize ±0.96 & 36.32  \scriptsize ±0.13 & -53.26  \scriptsize ±0.75 \\
- & LwF    & 19.61  \scriptsize ±0.05 & -96.69  \scriptsize ±0.25 & 63.29  \scriptsize ±2.35 & -32.56  \scriptsize ±0.56 & 8.46  \scriptsize ±0.22 & -76.74  \scriptsize ±0.44 & 15.85  \scriptsize ±0.58 & -67.79  \scriptsize ±0.23 \\ \hline
\multirow{10}{*}{200} & GEM                     & 25.54  \scriptsize ±0.76          & -82.61  \scriptsize ±1.60          & 90.44  \scriptsize ±0.94          & - 9.27  \scriptsize ±2.07          & -                     & -                      & -                     & -                      \\
                      & A-GEM                   & 20.04  \scriptsize ±0.34          & -95.73  \scriptsize ±0.20          & 83.88  \scriptsize ±1.49          & -16.39  \scriptsize ±0.80          &  8.07  \scriptsize ±0.08          & -77.02  \scriptsize ±0.22          & 22.77  \scriptsize ±0.03          & -56.61  \scriptsize ±0.32          \\
                      & iCaRL                   & 49.02  \scriptsize ±3.20          & -28.72  \scriptsize ±0.49          & 88.99  \scriptsize ±2.13          & - 1.01  \scriptsize ±4.15          &  7.53  \scriptsize ±0.79          & -22.70  \scriptsize ±0.44          & 28.19  \scriptsize ±1.47          & -10.36  \scriptsize ±0.31          \\
                      & GSS                     & 39.07  \scriptsize ±5.59          & -75.25  \scriptsize ±4.07          & 88.80  \scriptsize ±2.89          & - 8.56  \scriptsize ±1.78          & -                     & -                      & -                     & -                      \\ \cline{2-10} 
                      & ER                      & 55.84  \scriptsize ±0.71          & -47.77  \scriptsize ±1.39          & 92.41  \scriptsize ±0.59          & - 5.73  \scriptsize ±0.38          &  8.67  \scriptsize ±0.25          & -77.29  \scriptsize ±0.26          & 39.28  \scriptsize ±0.83          & -42.05  \scriptsize ±0.29          \\
                      & \cellcolor{gray!20} RER                             & \cellcolor{gray!20}\textbf{58.59  \scriptsize ±0.74}  & \cellcolor{gray!20}\textbf{-44.50  \scriptsize ±0.80} & \cellcolor{gray!20}\textbf{92.85  \scriptsize ±0.36}  & \cellcolor{gray!20}\textbf{- 5.40  \scriptsize ±0.65} & \cellcolor{gray!20}\textbf{ 9.35  \scriptsize ±0.21}  & \cellcolor{gray!20}\textbf{-76.67  \scriptsize ±0.38} & \cellcolor{gray!20}\textbf{40.83  \scriptsize ±0.58}  & \cellcolor{gray!20}\textbf{-41.19  \scriptsize ±0.55} \\ \cline{2-10} 
                      & ER-ACE                  & 63.02  \scriptsize ±1.29          & -20.35  \scriptsize ±1.76          & 92.59  \scriptsize ±0.36          & \textbf{- 5.33  \scriptsize ±0.41} & 11.67  \scriptsize ±0.29          & -49.03  \scriptsize ±1.61          & 42.08  \scriptsize ±0.35          & -37.71  \scriptsize ±1.10          \\
                      & \cellcolor{gray!20} RER-ACE                         & \cellcolor{gray!20}\textbf{63.52  \scriptsize ±0.71}  & \cellcolor{gray!20}\textbf{-20.11  \scriptsize ±5.66} & \cellcolor{gray!20}\textbf{92.63  \scriptsize ±0.58}  & \cellcolor{gray!20}- 5.43  \scriptsize ±0.67          & \cellcolor{gray!20}\textbf{12.18  \scriptsize ±0.41}  & \cellcolor{gray!20}\textbf{-48.91  \scriptsize ±2.59} & \cellcolor{gray!20}\textbf{44.11  \scriptsize ±0.62}  & \cellcolor{gray!20}\textbf{-36.62  \scriptsize ±1.17} \\ \cline{2-10} 
                      & DER++                   & 62.30  \scriptsize ±1.07          & -35.83  \scriptsize ±1.34          & 90.74  \scriptsize ±1.01          & - 7.45  \scriptsize ±1.07          & 12.26  \scriptsize ±0.31          & -68.37  \scriptsize ±1.38          & 40.47  \scriptsize ±1.53          & -40.41  \scriptsize ±1.29          \\
                      & \cellcolor{gray!20} RDER                            & \cellcolor{gray!20}\textbf{65.38  \scriptsize ±0.42}  & \cellcolor{gray!20}\textbf{-34.16  \scriptsize ±1.90} & \cellcolor{gray!20}\textbf{91.67  \scriptsize ±0.80}  & \cellcolor{gray!20}\textbf{- 6.81  \scriptsize ±1.19} & \cellcolor{gray!20}\textbf{13.96  \scriptsize ±0.64}  & \cellcolor{gray!20}\textbf{-67.02  \scriptsize ±1.24} & \cellcolor{gray!20}\textbf{40.87  \scriptsize ±0.92}  & \cellcolor{gray!20}\textbf{-39.87  \scriptsize ±1.31} \\ \hline
\multirow{10}{*}{500} & GEM                     & 26.20  \scriptsize ±1.26          & -74.31  \scriptsize ±4.62          & 92.16  \scriptsize ±0.69          & - 9.12  \scriptsize ±0.21          & -                     & -                      & -                     & -                      \\
                      & A-GEM                   & 22.67  \scriptsize ±0.57          & -94.01  \scriptsize ±1.16          & 89.48  \scriptsize ±1.45          & -14.26  \scriptsize ±4.18          &  8.06  \scriptsize ±0.04          & -77.06  \scriptsize ±0.41          & 25.33  \scriptsize ±0.49          & -55.68  \scriptsize ±1.01          \\
                      & iCaRL                   & 47.55  \scriptsize ±3.95          & -25.71  \scriptsize ±1.10          & 88.22  \scriptsize ±2.62          & - 1.06  \scriptsize ±4.21          &  9.38  \scriptsize ±1.53          & -20.89  \scriptsize ±0.23          & 31.55  \scriptsize ±3.27          & - 7.30  \scriptsize ±0.79          \\
                      & GSS                     & 49.73  \scriptsize ±4.78          & -62.88  \scriptsize ±2.67          & 91.02  \scriptsize ±1.57          & - 7.73  \scriptsize ±3.99          & -                     & -                      & -                     & -                      \\ \cline{2-10} 
                      & ER                      & 69.01  \scriptsize ±0.37          & -33.02  \scriptsize ±2.62          & 94.28  \scriptsize ±0.27          & \textbf{- 3.09  \scriptsize ±1.61} & 10.40  \scriptsize ±0.16          & -74.36  \scriptsize ±0.58          & 48.82  \scriptsize ±0.34          & -31.06  \scriptsize ±1.53          \\
                      & \cellcolor{gray!20} RER                             & \cellcolor{gray!20}\textbf{69.22  \scriptsize ±1.96}  & \cellcolor{gray!20}\textbf{-29.79  \scriptsize ±2.87} & \cellcolor{gray!20}\textbf{94.50  \scriptsize ±0.41}  & \cellcolor{gray!20}- 3.40  \scriptsize ±0.33          & \cellcolor{gray!20}\textbf{11.50  \scriptsize ±0.47}  & \cellcolor{gray!20}\textbf{-74.13  \scriptsize ±0.72} & \cellcolor{gray!20}\textbf{51.28  \scriptsize ±0.93}  & \cellcolor{gray!20}\textbf{-30.29  \scriptsize ±1.24} \\ \cline{2-10} 
                      & ER-ACE                  & 71.26  \scriptsize ±0.66          & -13.37  \scriptsize ±1.06          & \textbf{94.31  \scriptsize ±0.23} & - 3.19  \scriptsize ±0.39          & 19.59  \scriptsize ±0.13          & -47.56  \scriptsize ±0.68          & 50.99  \scriptsize ±0.45          & -29.32  \scriptsize ±0.46          \\
                      & \cellcolor{gray!20} RER-ACE                         & \cellcolor{gray!20}\textbf{71.29  \scriptsize ±1.15}  & \cellcolor{gray!20}\textbf{-12.53  \scriptsize ±2.41} & \cellcolor{gray!20}94.25  \scriptsize ±0.23           & \cellcolor{gray!20}\textbf{- 3.16  \scriptsize ±0.80} & \cellcolor{gray!20}\textbf{20.41  \scriptsize ±0.66}  & \cellcolor{gray!20}\textbf{-42.22  \scriptsize ±1.09} & \cellcolor{gray!20}\textbf{54.62  \scriptsize ±0.87}  & \cellcolor{gray!20}\textbf{-25.15  \scriptsize ±0.98} \\ \cline{2-10} 
                      & DER++                   & 72.11  \scriptsize ±1.41          & -23.40  \scriptsize ±1.32          & \textbf{94.21  \scriptsize ±0.32} & - 3.98  \scriptsize ±0.60          & 19.29  \scriptsize ±1.14          & -60.58  \scriptsize ±0.46          & 51.39  \scriptsize ±0.91          & -26.90  \scriptsize ±0.52          \\
                      & \cellcolor{gray!20} RDER                            & \cellcolor{gray!20}\textbf{73.99  \scriptsize ±1.03}  & \cellcolor{gray!20}\textbf{-22.86  \scriptsize ±1.76} & \cellcolor{gray!20}94.04  \scriptsize ±0.43           & \cellcolor{gray!20}\textbf{- 3.82  \scriptsize ±0.59} & \cellcolor{gray!20}\textbf{20.06  \scriptsize ±1.18}  & \cellcolor{gray!20}\textbf{-56.16  \scriptsize ±1.38} & \cellcolor{gray!20}\textbf{52.56  \scriptsize ±0.69}  & \cellcolor{gray!20}\textbf{-25.02  \scriptsize ±0.24} \\ \hline
\multirow{10}{*}{5120} & GEM                     & 25.26  \scriptsize ±3.46          &-75.27  \scriptsize ±4.41          & 95.55  \scriptsize ±0.02          & - 6.91  \scriptsize ±2.33          & -                     & -                      & -                     & -                      \\
                      & A-GEM                   & 21.99  \scriptsize ±2.29          & -84.49  \scriptsize ±3.08          & 90.10  \scriptsize ±2.09          & - 9.89  \scriptsize ±0.40          & 7.96  \scriptsize ±0.13           & -76.01  \scriptsize ±0.52          & 26.22  \scriptsize ±0.65          & -55.61  \scriptsize ±0.84          \\
                      & iCaRL                   & 55.07  \scriptsize ±1.55          & -24.94  \scriptsize ±0.14          & 92.23  \scriptsize ±0.84          & - 0.99  \scriptsize ±1.41          & 14.08  \scriptsize ±1.92          & -16.00  \scriptsize ±0.28          & 40.83  \scriptsize ±3.11          & - 2.60  \scriptsize ±0.35          \\
                      & GSS                     & 67.27  \scriptsize ±4.27          & -58.11  \scriptsize ±9.12          & 94.19  \scriptsize ±1.15          & - 6.38  \scriptsize ±1.71          & -                     & -                      & -                     & -                      \\ \cline{2-10} 
                      & ER                      & 83.30  \scriptsize ±0.50          & -13.79  \scriptsize ±1.40          & 96.95  \scriptsize ±0.15          & - 0.98  \scriptsize ±0.36          & 28.52  \scriptsize ±0.37          & -52.54  \scriptsize ±1.45          & 68.46  \scriptsize ±0.40          & \textbf{-10.13  \scriptsize ±0.20} \\
                      & \cellcolor{gray!20} RER                             & \cellcolor{gray!20}\textbf{83.53  \scriptsize ±0.55}  & \cellcolor{gray!20}\textbf{-12.13  \scriptsize ±1.29} & \cellcolor{gray!20}\textbf{96.98  \scriptsize ±0.17}  & \cellcolor{gray!20}\textbf{- 0.79  \scriptsize ±0.24} & \cellcolor{gray!20}\textbf{33.86  \scriptsize ±0.64}  & \cellcolor{gray!20}\textbf{-45.56  \scriptsize ±2.13} & \cellcolor{gray!20}\textbf{69.31  \scriptsize ±0.41}  & \cellcolor{gray!20}-10.68  \scriptsize ±0.25          \\ \cline{2-10} 
                      & ER-ACE                  & 82.98  \scriptsize ±0.38          & \textbf{- 3.99  \scriptsize ±0.61} & 96.76  \scriptsize ±0.08          & - 0.63  \scriptsize ±0.30          & \textbf{37.02  \scriptsize ±0.17} & \textbf{-33.29  \scriptsize ±1.03} & 68.69  \scriptsize ±0.19          & - 9.88  \scriptsize ±0.42          \\
                      & \cellcolor{gray!20} RER-ACE                         & \cellcolor{gray!20}\textbf{83.74  \scriptsize ±0.79}  & \cellcolor{gray!20} - 4.05  \scriptsize ±1.81         & \cellcolor{gray!20}\textbf{96.80  \scriptsize ±0.20}  & \cellcolor{gray!20}\textbf{- 0.57  \scriptsize ±0.27} & \cellcolor{gray!20}36.97  \scriptsize ±0.94           & \cellcolor{gray!20}-33.79  \scriptsize ±3.07          & \cellcolor{gray!20}\textbf{69.05  \scriptsize ±0.73}  & \cellcolor{gray!20}\textbf{- 9.51  \scriptsize ±0.79} \\ \cline{2-10} 
                      & DER++                   & 84.50  \scriptsize ±0.63          & - 9.79  \scriptsize ±0.34                   & 95.91  \scriptsize ±0.57          & - 1.57  \scriptsize ±0.25                 & 37.88  \scriptsize ±0.37          & -30.62  \scriptsize ±1.78          & 68.05  \scriptsize ±0.53          & - 8.80  \scriptsize ±0.24          \\
                      & \cellcolor{gray!20} RDER                            & \cellcolor{gray!20}\textbf{85.56  \scriptsize ±0.38}  & \cellcolor{gray!20}\textbf{- 8.81  \scriptsize ±0.71} & \cellcolor{gray!20}\textbf{96.21  \scriptsize ±0.22}  & \cellcolor{gray!20}\textbf{- 1.42  \scriptsize ±0.09} & \cellcolor{gray!20}\textbf{39.67  \scriptsize ±0.96}  & \cellcolor{gray!20}\textbf{-29.37  \scriptsize ±1.53} & \cellcolor{gray!20}\textbf{68.82  \scriptsize ±0.54}  & \cellcolor{gray!20}\textbf{- 8.03  \scriptsize ±0.46} \\ \hline
\end{tabular}
}
\end{table*}

\begin{table*}[t]
\caption{Comparison of ACC with three different baselines on the CIFAR-100. The results of our method are shown in gray cells and the better results are presented in \textbf{bold}.}
\label{tab:cifar100}
\centering
\resizebox{0.9\textwidth}{!}{
\begin{tabular}{cc|c
>{\columncolor{gray!20}}c |c
>{\columncolor{gray!20}}c |c
>{\columncolor{gray!20}}c }
\hline
Settings                   & Buffer Size & ER                   & RER                   & ER-ACE       & RER-ACE               & DER++                 & RDER                  \\ \hline
                           & 100           & 11.31  \scriptsize ±0.21          & \textbf{13.94  \scriptsize ±0.89} & 18.59  \scriptsize ±1.08 & \textbf{20.20  \scriptsize ±0.86} & 14.98  \scriptsize ±0.65          & \textbf{20.79  \scriptsize ±1.05} \\
                           & 200           & 14.78  \scriptsize ±0.40          & \textbf{16.40  \scriptsize ±0.53} & 25.14  \scriptsize ±1.83 & \textbf{26.64  \scriptsize ±0.29} & 24.17  \scriptsize ±1.37          & \textbf{30.65  \scriptsize ±0.76} \\
                           & 500           & 23.10  \scriptsize ±0.32          & \textbf{26.97  \scriptsize ±0.75} & 36.02  \scriptsize ±0.84 & \textbf{36.06  \scriptsize ±1.14} & 35.19  \scriptsize ±1.30          & \textbf{39.50  \scriptsize ±1.54} \\
\multirow{-4}{*}{Class-IL} & 5120          & 51.43  \scriptsize ±1.01          & \textbf{54.08  \scriptsize ±0.63} & 53.93  \scriptsize ±2.04 & \textbf{54.38  \scriptsize ±1.08} & 55.58  \scriptsize ±1.86          & \textbf{60.07  \scriptsize ±0.23} \\ \hline
                           & 100           & 58.64  \scriptsize ±1.31          & \textbf{59.77  \scriptsize ±0.03} & 59.91  \scriptsize ±1.01 & \textbf{61.10  \scriptsize ±0.93} & 58.32  \scriptsize ±1.27          & \textbf{59.07  \scriptsize ±0.73} \\
                           & 200           & 66.31  \scriptsize ±0.76          & \textbf{66.83  \scriptsize ±0.97} & 64.81  \scriptsize ±3.14 & \textbf{67.42  \scriptsize ±0.60} & 66.47  \scriptsize ±0.57          & \textbf{68.60  \scriptsize ±0.51} \\
                           & 500           & 73.10  \scriptsize ±0.99           & \textbf{73.99  \scriptsize ±0.51} & 74.13  \scriptsize ±0.84 & \textbf{74.54  \scriptsize ±1.37} & 74.10  \scriptsize ±1.62          & \textbf{75.59  \scriptsize ±0.85} \\
\multirow{-4}{*}{Task-IL}  & 5120          & \textbf{86.16  \scriptsize ±0.47} & 85.35  \scriptsize ±0.34          & 84.69  \scriptsize ±1.32 & \textbf{85.15  \scriptsize ±0.41} & 86.23  \scriptsize ±2.18          & \textbf{86.54  \scriptsize ±0.31} \\ \hline
\end{tabular}}
\vspace{-2mm}
\end{table*}

\subsection{Comparison with State-of-the-art Methods}
\label{subsec:exp-results}

As previously mentioned, our proposed method aims to enhance general \textit{rehearsal-based} baselines by taking into account the `stability-plasticity' dilemma, which involves the relationship between new and old tasks, and the specific importance of different samples. We integrate this approach into three representative baselines: ER, ER-ACE~\cite{ER-ACE}, and DER++~\cite{DER}, which are termed Relational-ER (RER), Relational-ER-ACE (RER-ACE), and Relational-DER (RDER), respectively. Implementation details can be found in Appendix~\ref{appendix:apply-other-baselines}.

Table~\ref{tab:baseline} presents the comparison results between our proposed algorithm applied on the three baselines and various state-of-the-art methods over ACC and BWT metrics. These comparison methods include three \textit{regularization-based} methods: oEWC~\cite{oEWC}, SI~\cite{SI}, and LwF~\cite{LwF}, and four \textit{rehearsal-based} methods: GEM~\cite{GEM}, A-GEM~\cite{A-GEM}, iCaRL~\cite{icarl}, and GSS~\cite{GSS}.
Besides, we also provide an upper-bound method and a lower-bound method for better reference, where the former trained on all data from old and new tasks together and the latter directly trained on the new task without any strategies to prevent model forgetting.

For CIFAR-10, it can be observed that our proposed approach can be adapted to different rehearsal-based baselines and achieve consistent performance improvement.
For example, the proposed RDER achieves as much as 3.08\% absolute performance gain compared to the baseline DER++ with a memory buffer of 200 under Class-IL.
Besides, our method achieves significantly higher classification accuracy than all comparison state-of-the-art methods under different settings.
On the other hand, our method can also significantly reduce the BWT of baselines, indicating that the proposed method can effectively reduce model forgetting while improving classification accuracy, \textit{i.e.}, achieving a better balance of `stability' and `plasticity'.

For Tiny-ImageNet dataset, our method also consistently achieves the best results under almost all settings.
Although iCaRL achieves sound BWT metric, it falls short in terms of ACC, revealing that a model that pays much attention to avoid forgetting may negatively impact classification performance. In contrast, our method improves both ACC and BWT compared to the corresponding baselines, highlighting that our method can achieve a better trade-off between new and old tasks.
Notably, the results of GEM and GSS are not reported in Table~\ref{tab:baseline} since the excessive computational overhead is unacceptable. 

Moreover, we validate the effectiveness of our method on CIFAR-100 in Table~\ref{tab:cifar100}. Obviously, our method can achieve a significant improvement for all these three baselines across different buffer sizes. For instance, in the Class-IL setting with a buffer size of 100, our method applied to ER, ER-ACE, and DER can improve their ACC by 2.63\%, 1.61\%, and 5.81\%, respectively. These results further demonstrate the strong adaptability of our method to multiple datasets and diverse settings.

\begin{figure*}[!t]
\centering
\subfloat[\tiny CIFAR-10 ER/RER]{\includegraphics[width=0.28\textwidth]{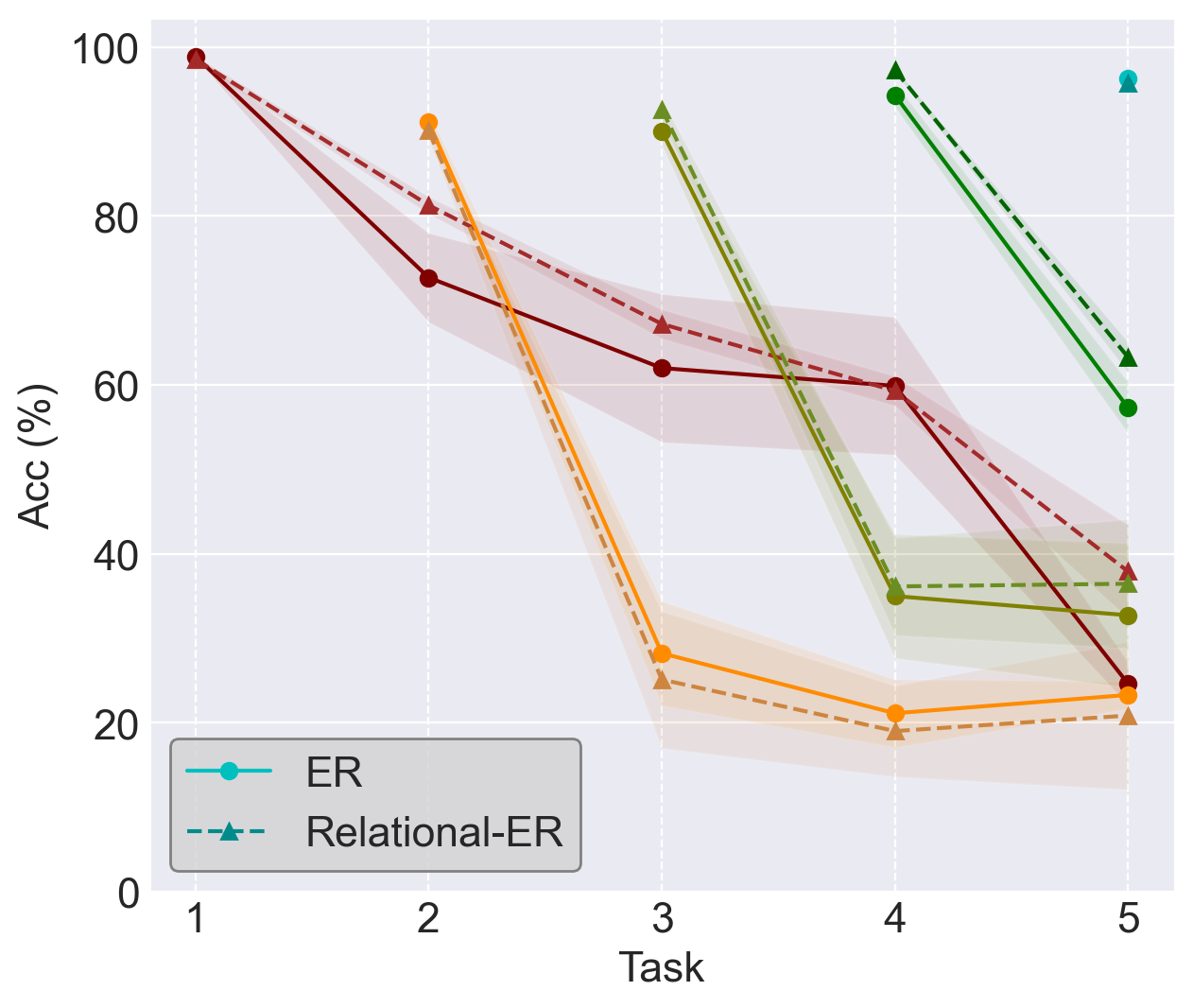}
\label{subfig:cifar10-er}}
\hfil
\subfloat[\tiny CIFAR-10 ER-ACE/RER-ACE]{\includegraphics[width=0.28\textwidth]{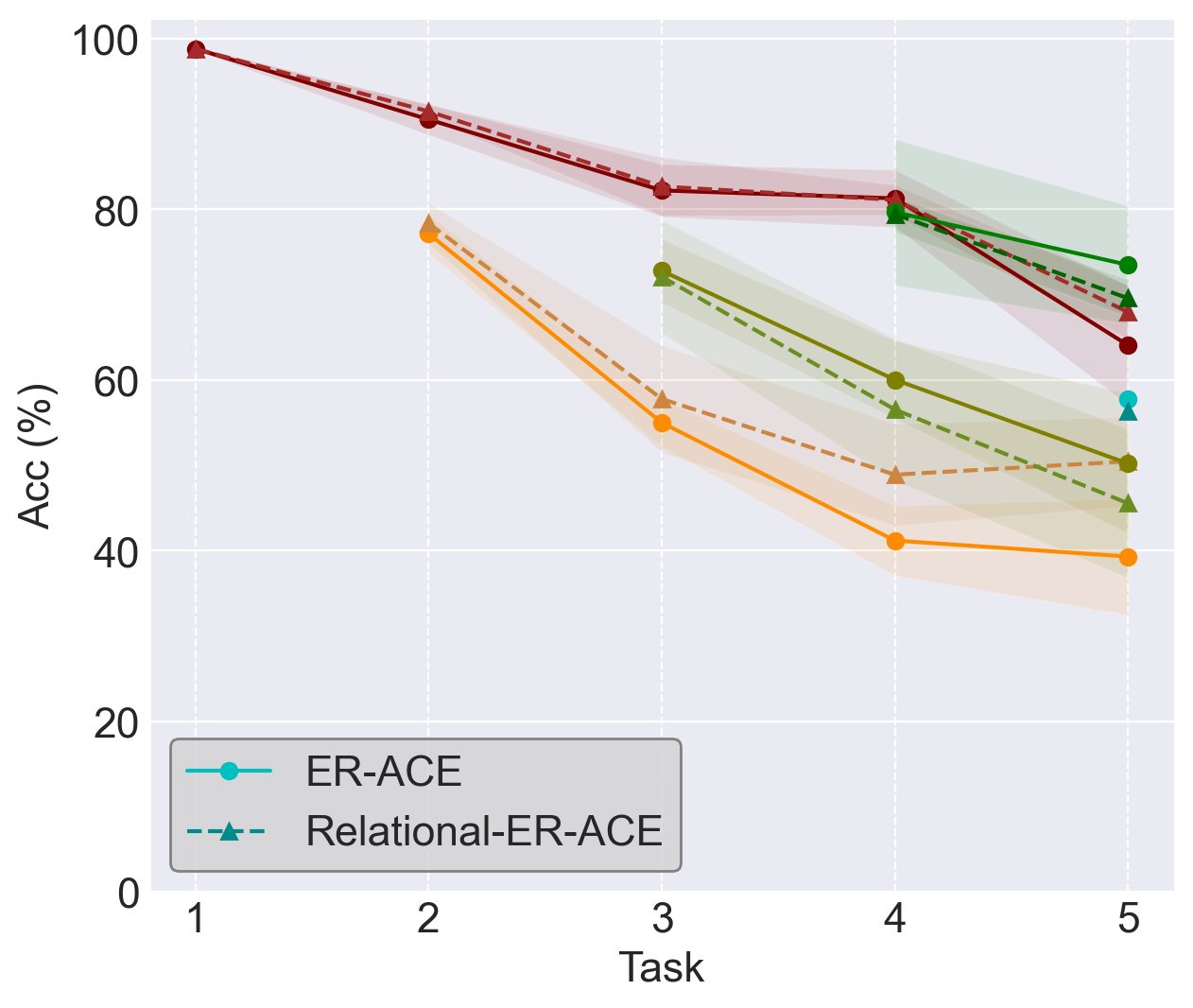}%
\label{subfig:cifar10-er-ace}}
\hfil
\subfloat[\tiny CIFAR-10 DER++/RDER]{\includegraphics[width=0.28\textwidth]{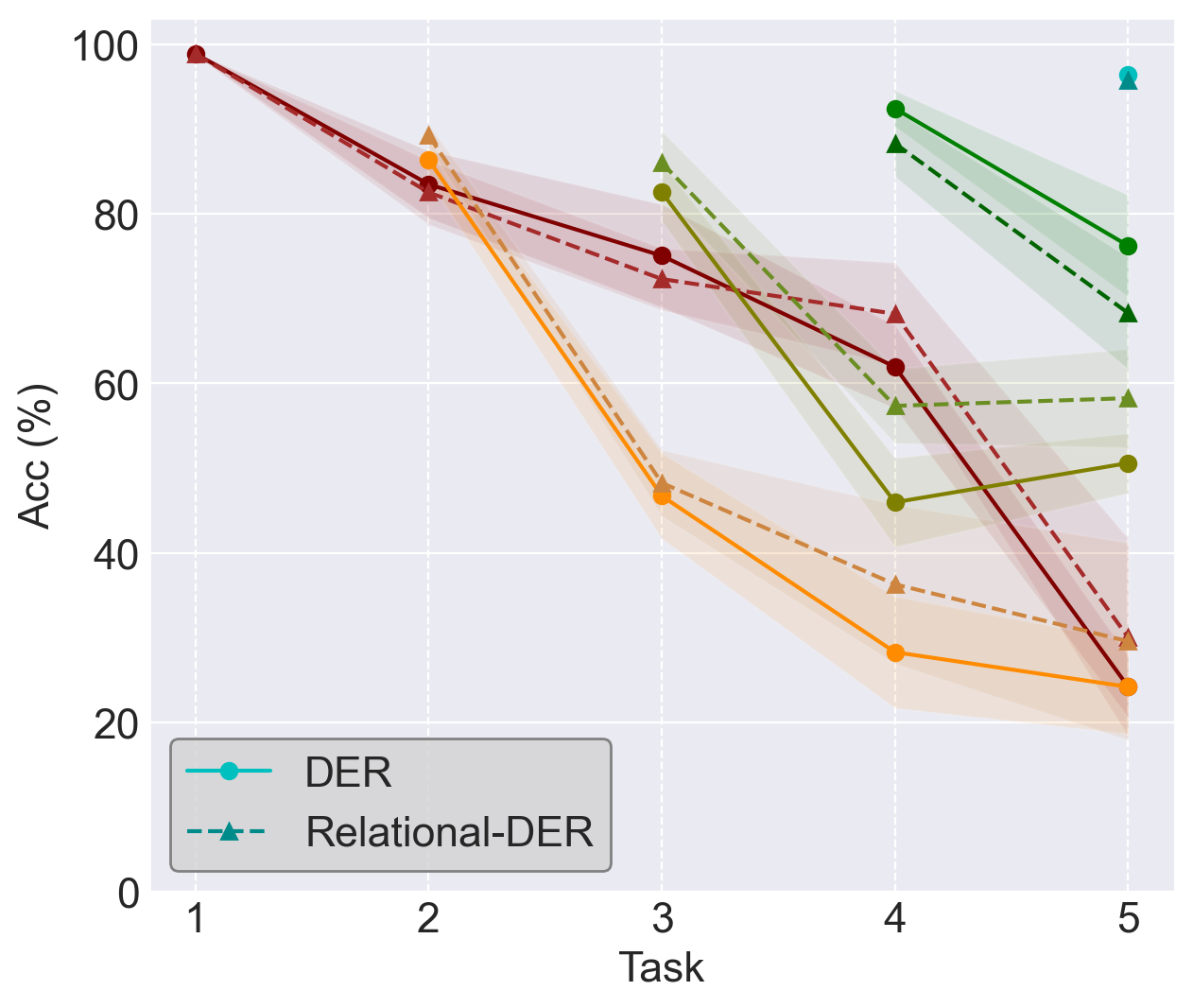}%
\label{subfig:cifar10-der}}

\subfloat[\tiny CIFAR-100 ER/RER]{\includegraphics[width=0.44\textwidth]{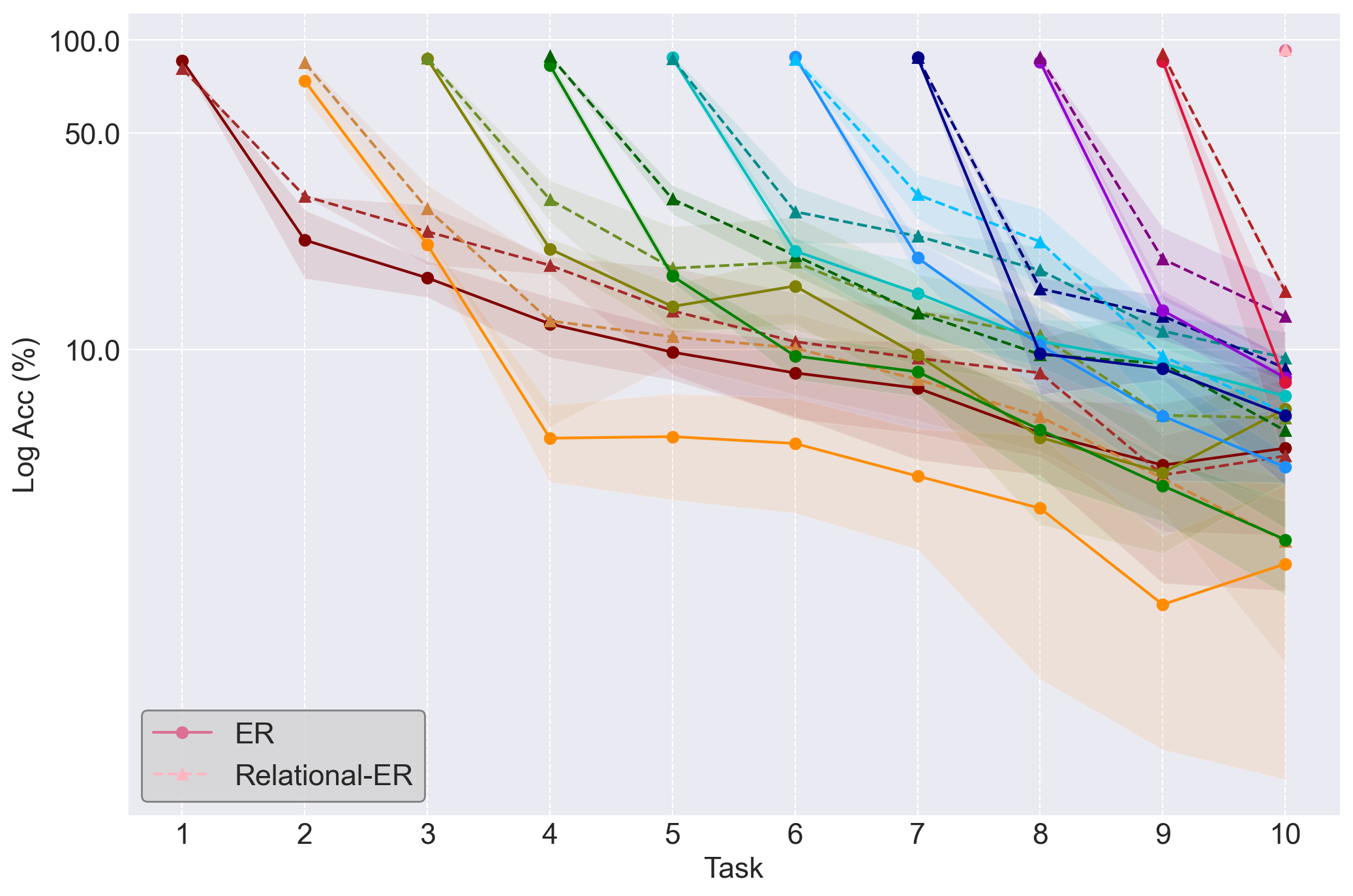}%
\label{subfig:cifar100-er}}
\hfil
\subfloat[\tiny Tiny-ImageNet ER/RER]{\includegraphics[width=0.44\textwidth]{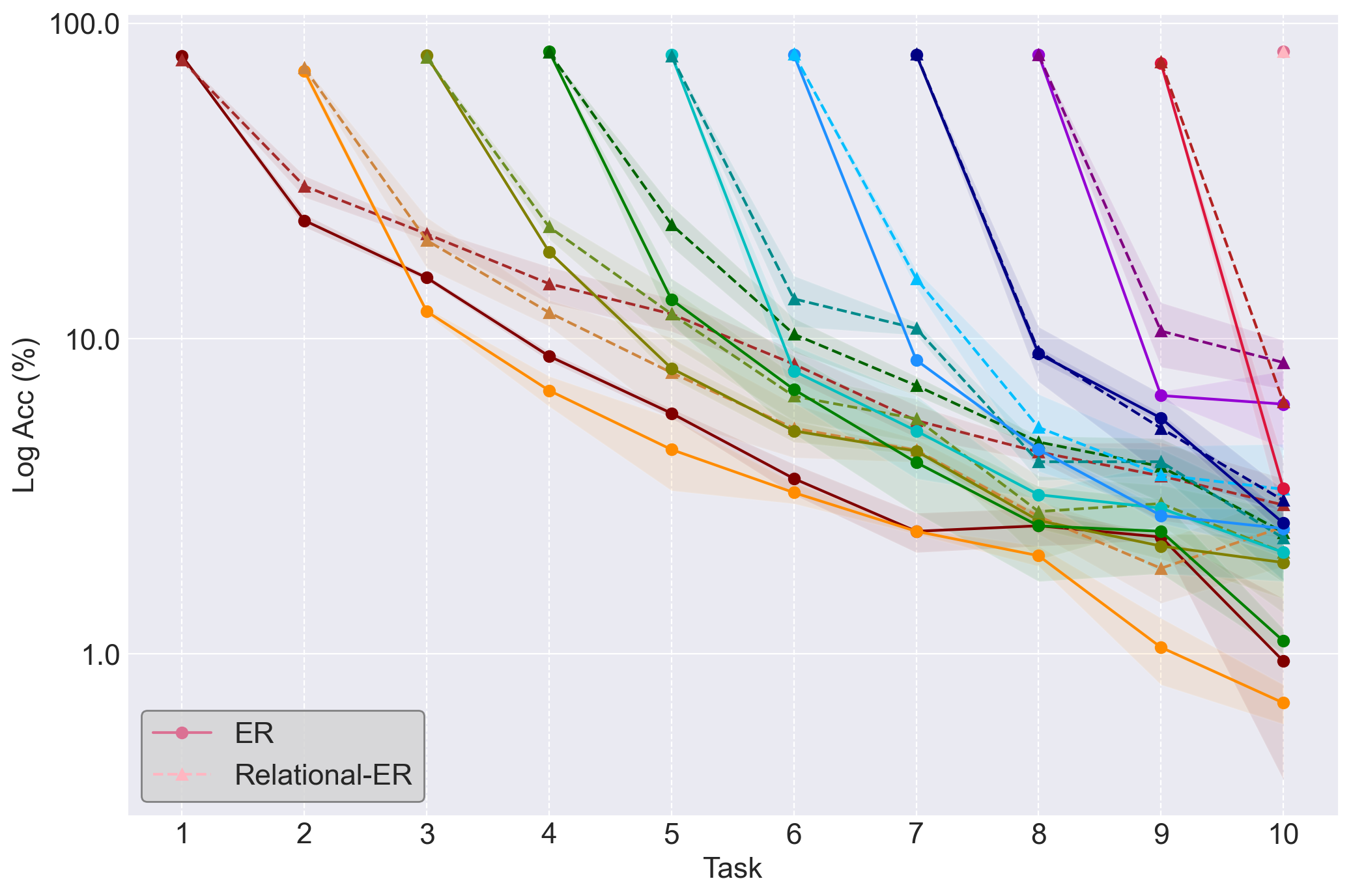}%
\label{subfig:tinyimg-er}}
\caption{Classification accuracy (\%) of each task during the whole training process. The dot-solid lines represent the comparison methods ER, ER-ACE, or DER++ and the triangle-dashed lines represent our methods RER, RER-ACE, or RDER.}
\label{fig:acc_per_task}
\end{figure*}

\subsection{Discussion}
\label{subsec:discussion}
To further explore the proposed method, we conduct more experiments and analyses in this section.

\medskip
\noindent\textbf{Does Our Method Mitigate the Model Forgetting?}
In order to better analyze the forgetting problem for old tasks, we visualize the accuracy change for each task during the continual learning process in Fig.~\ref{fig:acc_per_task} and Appendix~\ref{appendix:visualization}. The visualization results show that our method better mitigates forgetting for previous tasks, \textit{i.e.}, consolidating knowledge for old tasks better than corresponding baselines. Note that the performance of RDER is slightly lower than DER++ for some new tasks (shown in Fig.~\ref{subfig:cifar10-der}). This is because our method aims to improve the generalization of all tasks, not just pay attention to new tasks with more data.

\medskip
\noindent\textbf{How Task Similarity Affects Continual Learning Models?}
In this section, we further investigate the impact of task similarity on the generation of sample weights. From CIFAR-10, we choose two classes belonging to the `object' category, namely `ship' and `truck', as task 1. For task 2, we consider two distinct setups: \textbf{Setup 1}, which includes `airplane' and `automobile', and \textbf{Setup 2}, which includes `cat' and `horse'. Notably, \textbf{Setup 1} is a relatively semantic-relevant task since its classes also belong to the `object' category. Conversely, both classes in \textbf{Setup 2} belong to the `animals' category, rendering it a relatively semantic-irrelevant task with lower similarity to task 1.
Here we focus on the baseline ER and our proposed RER, because ER employs the same loss function (\textit{i.e.}, CE loss) for both new and old tasks, and the corresponding sample weights directly reflect their respective contributions to the training of the Main Net\footnote{Here we exclude DER++ and ER-ACE from our analysis since they adopt different loss functions and may introduce additional sources of interference.}.

\begin{figure}
    \centering
    \begin{minipage}[b]{0.45\textwidth}
        \centering
        \small
        \resizebox{0.9\textwidth}{!}{
        \begin{tabular}{cccc}
            \hline
            Method  & Setup 1          & Setup 2           & Gap\\
            \hline
            ER      & 81.38 $\pm$ 1.67 & 88.29 $\pm$ 0.89 & 6.91\\
            RER     & 84.32 $\pm$ 0.81 & 89.41 $\pm$ 1.60 & 5.09\\
            \hline
        \end{tabular}}
        \label{fig:similar-task-acc}
        \caption*{(a) ACC of the ER and RER under \textbf{Setup 1} (semantic-relevant task) and \textbf{Setup 2} (semantic-irrelevant task). The gap in the last column indicates the difference between the ACC of \textbf{Setup 1} and \textbf{Setup 2}.}
    \end{minipage}

    \begin{minipage}[b]{0.45\textwidth}
        \centering
        \includegraphics[width=0.75\textwidth]{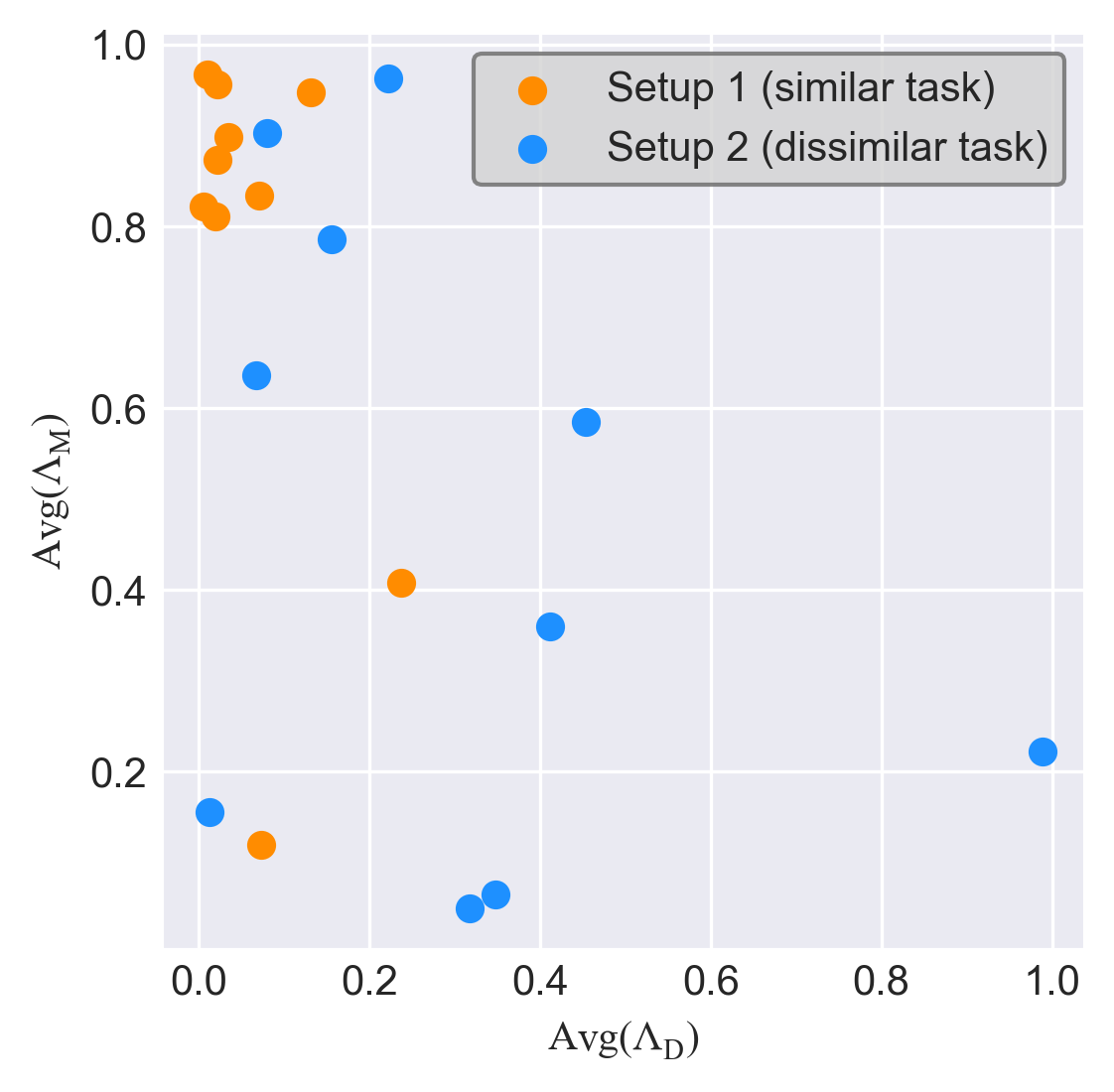}
        \label{fig:similar-task-weight}
        \vspace{-3mm}
        \caption*{(b) Visualization of the average sample weights of new task (horizontal axis) and old task (vertical axis) of \textbf{Setup 1} and \textbf{Setup 2}.}
    \end{minipage}
\caption{The visualization of task similarity effects.}
\label{fig:similar-task}
\end{figure}

To mitigate the impact of random factors, we trained ER/RER 10 times and report the ACC under \textbf{Setup 1} and \textbf{Setup 2} in Fig.~\ref{fig:similar-task}(a). The ACC of \textbf{Setup 1} is significantly lower than that of \textbf{Setup 2} for both ER and RER, implying that the extracted knowledge of the new task in \textbf{Setup 1} is similar to the old task, which may interfere with the classification of the old task. Moreover, our proposed RER exhibits a smaller performance gap between \textbf{Setup 2} and \textbf{Setup 1} than ER, indicating that our method can effectively extract the task relationship and mitigates interference from new tasks and resulting in overall performance enhancement.

Furthermore, at the end of task 2 in RER, we calculate the average weights for validation samples from the old and new tasks, respectively. Fig.~\ref{fig:similar-task}(b) shows the results for 10 different runs of RER under \textbf{Setup 1} (represented by `dark orange' scatters) and \textbf{Setup 2} (represented by `dodgerblue' scatters).
It is clear that under \textbf{Setup 1}, the RRN tends to generate larger weights for the old task samples and smaller weights for the new task samples. This suggests that for a similar new task, the old task is more important in ameliorating the forgetting because similar tasks may generate similar features which will interfere with each other.
On the other hand, under \textbf{Setup 2}, the average sample weights of the old and new tasks are more dispersed between 0 and 1, indicating that dissimilar (\textit{i.e.}, semantic-irrelevant) new tasks may not strongly interfere with the old tasks and the sample weights for each task should be determined in real-time during the whole training process.


\medskip
\noindent\textbf{Why We Need the Bi-level Optimization?}
To evaluate the effectiveness of the bi-level optimization paradigm, we conduct an ablation study by end-to-end training of the Main Net and the RRN together based on DER++ (referred to as Vanilla). Specifically, we employ the same RRN to generate the paired sample weights and update the Main Net and the RRN by a single backward step through $\mathcal{L}^{tr} + \mathcal{L}^{bf}$. In Table~\ref{table:cifar10_vanilla}, we observe that the sample weights generated by the Vanilla method appear to be uninformative, and even impair the performance of the baseline approach (DER++), indicating that the end-to-end training paradigm cannot produce meaningful weights to enhance the generalization capability of the Main Net. In contrast, our RDER approach consistently outperforms both DER++ and the Vanilla method, thereby demonstrating the effectiveness of the proposed bi-level optimization framework to achieve a better trade-off between `stability' and `plasticity'.

\begin{table}[t]
    \centering
    \caption{ACC on the CIFAR-10 dataset of the vanilla training and our proposed bi-level optimization framework.}
    \begin{tabular}{cccc}
        \hline
        \textbf{Memory} & \multirow{2}*{\textbf{Method}} & \multirow{2}*{\textbf{Class-IL}} & \multirow{2}*{\textbf{Task-IL}}\\
        \textbf{Size}& & & \\
        \hline
        \multirow{3}*{200}
          & DER++~\cite{DER}       & 62.30 $\pm$ 1.07 & 90.74 $\pm$ 1.01\\
          & RDER (ours) & \textbf{65.38} $\pm$ 0.42 & \textbf{91.67} $\pm$ 0.80\\
          & Vanilla     & 62.85 $\pm$ 2.90 & 91.20 $\pm$ 1.88\\
        \hline
        \multirow{3}*{500}
          & DER++~\cite{DER}       & 72.11 $\pm$ 1.41 & 94.21 $\pm$ 0.32\\
          & RDER (ours) & \textbf{73.99} $\pm$ 1.03 & \textbf{94.04} $\pm$ 0.43\\
          & Vanilla     & 72.28 $\pm$ 0.93 & 93.42 $\pm$ 0.75\\
        \hline
        \multirow{3}*{5120}
          & DER++~\cite{DER}       & 84.50 $\pm$ 0.63 & 95.91 $\pm$ 0.57\\
          & RDER (ours) & \textbf{85.56} $\pm$ 0.38 & \textbf{96.21} $\pm$ 0.22\\
          & Vanilla     & 82.26 $\pm$ 2.45 & 95.3 $\pm$ 0.72\\
        \hline
    \end{tabular}
    \label{table:cifar10_vanilla}
\end{table}

\begin{table}[t]
    \centering
    \caption{ACC on the CIFAR-10 dataset of different ways to construct the training set in the outer loop.}
    \begin{tabular}{cccc}
        \hline
        \textbf{Memory} & \multirow{2}*{\textbf{Method}} & \multirow{2}*{\textbf{Class-IL}} & \multirow{2}*{\textbf{Task-IL}}\\
        \textbf{Size}& & & \\
        \hline
        \multirow{3}*{200}
          & DER++~\cite{DER}       & 62.30 $\pm$ 1.07          & 90.74 $\pm$ 1.01         \\
          & RDER (ours) & \textbf{65.38} $\pm$ 0.42 & \textbf{91.67} $\pm$ 0.80\\
          & Split RDER  & 62.53 $\pm$ 0.66          & 91.36 $\pm$ 0.77         \\
        \hline
        \multirow{3}*{500}
          & DER++~\cite{DER}       & 73.11 $\pm$ 1.41          & 94.21 $\pm$ 0.32         \\
          & RDER (ours) & \textbf{73.99} $\pm$ 1.03 & \textbf{94.04} $\pm$ 0.43         \\
          & Split RDER  & 72.57 $\pm$ 0.73          & 93.77 $\pm$ 0.33\\
        \hline
        \multirow{3}*{5120}
          & DER++~\cite{DER}       & 84.50 $\pm$ 0.63          & 95.91 $\pm$ 0.57         \\
          & RDER (ours) & \textbf{85.56} $\pm$ 0.38 & \textbf{96.21} $\pm$ 0.22\\
          & Split RDER  & 85.35 $\pm$ 0.24          & 96.19 $\pm$ 0.46         \\
        \hline
    \end{tabular}
    \label{table:meta_set}
\end{table}

\medskip
\noindent\textbf{Why Use the Memory Buffer to Train Relation Replay Net in the Outer Loop?}
In Eq.~(\ref{eq:outer-loss}) of the outer-loop optimization, we utilize the memory buffer $\mathcal{M}$ to train the RRN. However, concerns arise regarding the possibility of overfitting when using the memory buffer to guide the RRN. To address this, we investigate two approaches for constructing the outer-loop training set: 1) Splitting the memory buffer $\mathcal{M}$ into two sets, which are utilized for training the Main Net in the inner loop and the RRN in the outer loop, respectively \cite{ICML2021_contextual}. 2) Alternatively, training the RRN in the outer loop using the entire memory buffer $\mathcal{M}$, as we propose.

We present a comparison between the two approaches based on DER++ and report the results in Table~\ref{table:meta_set}, where the two methods are denoted as `Split RDER' and `RDER', respectively. Split RDER divides the memory buffer into two sets for outer- and inner-loop training with a ratio of $20\%-80\%$, following \cite{ICML2021_contextual}, whereas the RDER approach does not involve such a split. The results in Table~\ref{table:meta_set} indicate that Split RDER exhibits a slight improvement over the baseline DER++, but its performance is still lower than our proposed RDER. These findings suggest that the first approach may lead to better generalization of the RRN while reducing the number of memory samples used to train the Main Net. However, this reduction in training samples may create a more serious imbalance between the new and old tasks, thereby reducing the overall performance, particularly when the buffer size is limited.

\subsection{Ablation Study}
\label{subsec:ablation}
In this section, we first validate the effect of the small buffer sizes on our method based on the three baselines. And then we conduct a detailed ablation study based on RDER to evaluate the influence caused by two critical hyperparameters ${Iter}_{warm}$ and $Interval$ on CIFAR-10 with different buffer sizes.

\begin{table*}[t]
\caption{Comparison of ACC on the CIFAR-10 with small buffer sizes. The results of our method are shown in gray cells and the better results are presented in \textbf{bold}.}
\label{tab:cifar10-small}
\centering
\resizebox{0.9\textwidth}{!}{
\begin{tabular}{cc|c
>{\columncolor{gray!20}}c |c
>{\columncolor{gray!20}}c |c
>{\columncolor{gray!20}}c }
\hline
Settings                   & Buffer Size & ER                    & RER                   & ER-ACE                & RER-ACE               & DER++        & RDER                  \\ \hline
                           & 50            & 36.47  \scriptsize ±2.92          & \textbf{37.43  \scriptsize ±1.27} & \textbf{42.16  \scriptsize ±1.78} & 41.95  \scriptsize ±1.07          & 46.70  \scriptsize ±4.14 & \textbf{49.75  \scriptsize ±1.92} \\
\multirow{-2}{*}{Class-IL} & 100           & 46.75  \scriptsize ±1.58          & \textbf{50.85  \scriptsize ±1.14} & 56.96  \scriptsize ±2.39          & \textbf{57.98  \scriptsize ±3.02} & 53.82  \scriptsize ±1.21 & \textbf{56.36  \scriptsize ±1.54} \\ \hline
                           & 50            & \textbf{88.13  \scriptsize ±1.16} & 88.07  \scriptsize ±0.80          & \textbf{87.09  \scriptsize ±1.05} & 86.45  \scriptsize ±1.05          & 84.00  \scriptsize ±0.84 & \textbf{85.81  \scriptsize ±1.99} \\
\multirow{-2}{*}{Task-IL}  & 100           & 90.27  \scriptsize ±1.05          & \textbf{90.56  \scriptsize ±0.41} & 90.28  \scriptsize ±0.38          & \textbf{91.01  \scriptsize ±0.04} & 87.48  \scriptsize ±1.43 & \textbf{89.16  \scriptsize ±1.37} \\ \hline
\end{tabular}}
\end{table*}

\medskip
\noindent\textbf{Effect of Small Buffer Size.}
To further investigate the impact of small buffer sizes on our method, we evaluate our method applied to the three baselines on CIFAR-10 in Table~\ref{tab:cifar10-small}. Specifically, RDER enhances DER++ by about +3.05\% with a buffer size of 50 under the Class-IL setting, demonstrating that our method can further explore the information in the memory buffer to enhance the Main Net overall performance.

\medskip
\noindent\textbf{Impact of the Warm-up Stage (${Iter}_{warm}$).}
Table~\ref{tab:iter-meta} presents the evaluation results for the various ${Iter}_{warm}$ under three different settings $\left[\frac{1}{3}, \frac{1}{2}, \frac{2}{3} \right]\times Iter_{max}$. The performance of RDER deteriorates significantly when a small value of ${Iter}_{warm}$ (\textit{i.e.}, 17) is used, suggesting that the RRN requires an adequate number of warm-up steps to generate meaningful sample weights to accurately capture the task-wise relationship and sample importance within each task. On the other hand, it can be observed a slight drop in performance under ${Iter}_{warm}=33$, since the preset weights may mislead the training in the warm-up stage. An excessively long warm-up stage also leads to insufficient iterations for the Main Net training guided by the generated weights.
Hence, we recommend setting ${Iter}_{warm}$ to be half of the number of iteration epochs for each task, which performs the best in Table~\ref{tab:iter-meta}.

\medskip
\noindent\textbf{Impact of the Relation Replay Net Updating Interval ($Interval$).}
Here we vary the value of $Interval$ to investigate its impact on our framework. As shown in Table~\ref{tab:interval}, the setting $Interval=5$ yields the highest classification accuracy across a range of memory buffer sizes.
However, small $Interval$, such as $Interval=1$, often results in decreased performance due to the frequent alternation between updating the Main Net and the RRN, leading to oscillation during training. Additionally, frequent updates of the RRN are computationally expensive, which can impede the convergence of the overall framework. Conversely, large $Interval$, such as $Interval=10$, can lead to faster computation, but we observed a significant drop in performance due to inadequate training of the RRN, which can result in the generation of suboptimal sample weights.
To strike a balance between accuracy and computational efficiency, we propose an empirical formulation $Interval = \#{epoch} / 10$, where $\#{epoch}$ represents the number of iteration epochs for each task.

\begin{table}[]
\caption{ACC on CIFAR-10 with different length of warm-up stage (${Iter}_{warm}$).}
\label{tab:iter-meta}
\begin{center}
    \begin{tabular}{cccc}
        \hline
        \textbf{\tabincell{c}{Memory\\Size}} & \textbf{${Iter}_{warm}$} & \textbf{Class-IL} & \textbf{Task-IL} \\
        \hline
        \multirow{3}{*}{50}  & 17         & 48.02  \scriptsize ±0.70          & 85.13  \scriptsize ±1.27          \\
                             & 25         & \textbf{49.75  \scriptsize ±1.92} & \textbf{85.81  \scriptsize ±1.99} \\
                             & 33         & 47.41  \scriptsize ±0.91          & 85.12  \scriptsize ±1.35          \\
        \hline
        \multirow{3}{*}{200} & 17         & 64.53  \scriptsize ±1.13          & 91.84  \scriptsize ±0.29          \\
                             & 25         & \textbf{65.38  \scriptsize ±0.42} & 91.67  \scriptsize ±0.80          \\
                             & 33         & 64.84  \scriptsize ±1.08          & \textbf{92.88  \scriptsize ±0.42} \\
        \hline
        \multirow{3}{*}{500} & 17         & 72.17  \scriptsize ±1.11          & 93.57  \scriptsize ±0.17          \\
                             & 25         & \textbf{73.99  \scriptsize ±1.03} & \textbf{94.04  \scriptsize ±0.43} \\
                             & 33         & 73.03  \scriptsize ±1.49          & 93.85  \scriptsize ±0.44          \\
        \hline
    \end{tabular}
\end{center}
\vspace{-5mm}
\end{table}

\begin{table}[]
\caption{ACC on CIFAR-10 with different values of Relation Replay Net updating interval ($Interval$).}
\label{tab:interval}
\begin{center}
    \begin{tabular}{cccc}
        \hline
        \textbf{\tabincell{c}{Memory\\Size}} & \textbf{$Interval$} & \textbf{Class-IL} & \textbf{Task-IL} \\
        \hline
        \multirow{3}{*}{50}  & 1        & 48.12  \scriptsize ±0.51          & 84.44  \scriptsize ±1.48          \\
                             & 5        & \textbf{49.75  \scriptsize ±1.92} & \textbf{85.81  \scriptsize ±1.99} \\
                             & 10       & 46.21  \scriptsize ±1.00          & 85.18  \scriptsize ±1.65          \\
        \hline
        \multirow{3}{*}{200} & 1        & 64.27  \scriptsize ±1.46          & 91.57  \scriptsize ±0.59          \\
                             & 5        & \textbf{65.38  \scriptsize ±0.42} & 91.67  \scriptsize ±0.80          \\
                             & 10       & 64.66  \scriptsize ±0.29          & \textbf{91.71  \scriptsize ±0.62} \\
        \hline
        \multirow{3}{*}{500} & 1        & 71.23  \scriptsize ±0.99          & 93.15  \scriptsize ±0.68          \\
                             & 5        & \textbf{73.99  \scriptsize ±1.03} & \textbf{94.04  \scriptsize ±0.43} \\
                             & 10       & 72.11  \scriptsize ±0.46          & 93.62  \scriptsize ±0.23          \\
        \hline
    \end{tabular}
\end{center}
\vspace{-5mm}
\end{table}

\section{Conclusion}
\label{sec:conclusion}
In this paper, we focus on the `stability-plasticity' dilemma in continual learning and strive to adaptively tune the relationship across different tasks and samples. To this end, we propose a novel continual learning framework, Relational Experience Replay, which pair-wisely adjusts the sample weights of samples from new tasks and the memory buffer. The sample weights generated by the Relation Replay Net can facilitate the optimization of the Main Net to achieve a better trade-off between `stability' and `plasticity'.
The proposed method can be easily integrated with multiple \textit{rehearsal-based} methods to achieve significant improvements. We theoretically and experimentally verify that the generated sample weights can extract the relationship between new and old tasks to automatically adjust the Main Net training and enhance the overall performance.
We expect that our method can provide more insights into the `stability-plasticity' dilemma and promote the development of the field of continual learning.



{
\appendices
\section{Calculation Details about the Relation Replay Net Updating}
\label{appendix:update_meta_net}
In this section, we provide a detailed calculation of the derivatives of the RRN. Referring back to Section~\ref{sec:update_meta_net}, the gradient descent of the RRN parameters is given in Eq. (\ref{eq:outer-update}) and repeated here as
\begin{equation}
    \phi^{k+1} = \phi^{k} - 
    \eta_\phi \triangledown_\phi \mathcal{L}^{bf}(\mathcal{B}^{bf}; \theta(\phi)),
\end{equation}
where the $\theta(\phi)$ is the one-step updated parameters generated by Eq.~(\ref{eq:inner-update}).

Here we can use the chain rule to calculate the derivative of $\phi$ as follows.
\begin{sequation}
\begin{split}
    &\triangledown_\phi \mathcal{L}^{bf} (\mathcal{B}^{bf}; \theta^k(\phi)) \\
    = &\frac{1}{B} \sum_{i=1}^{B} \triangledown_\phi L^{bf}(x_i; \theta^k(\phi)) \\
    = &\frac{1}{B}
    \sum_{i=1}^{B}
    \left. \frac{\partial L^{bf}(x_i; \theta)}{\partial \theta} \right |_{\theta^k} 
    \left. \frac{\partial \theta(\phi)}{\partial \phi} \right |_{\phi^{k}},   
\end{split}
\label{eq:derivation_term1}
\end{sequation}
where the second term can be represented as 
\begin{sequation}
\begin{split}
    \left. \frac{\partial \theta(\phi)}{\partial \phi} \right |_{\phi^{k}} &= 
        - \frac{\eta_\theta}{B} \sum_{j=1}^{B} 
        \left(
            \triangledown_\theta L^{tr}(x^D_j; \theta^k) 
            \cdot 
            \left. \frac{\partial \lambda^D_j(\phi)}{\partial \phi} \right |_{\phi^{k}} 
        \right. \\
        &\hspace{15mm} + \left.
            \triangledown_\theta L^{tr}(x^M_j; \theta^k)
            \cdot 
            \left. \frac{\partial \lambda^M_j(\phi)}{\partial \phi} \right |_{\phi^{k}}
        \right).
\end{split}
\label{eq:derivation_term2}
\end{sequation}
Then substituting Eq.~(\ref{eq:derivation_term2}) into Eq.~(\ref{eq:derivation_term1}) and exchanging the order of the two summations, we can get 
\begin{sequation}
\begin{split}
    &\hspace{5mm} \triangledown_\phi \mathcal{L}^{bf} (\mathcal{B}^{bf}; \theta^k(\phi)) \\
    &= \!- \frac{\eta_\theta}{B}\! \sum_{j=1}^{B}
    \!\left(
            \frac{1}{B} \sum_{i=1}^{B}
            \left. 
                \frac{\partial L^{bf}(x_i; \theta)}{\partial \theta} 
            \right|_{\theta^k}
            \!\!\! \cdot \!\!
            \left. 
                \frac{\partial L^{tr}(x^D_j; \theta)}{\partial \theta}
            \right|_{\theta^{k}}
        \!\!\! \cdot \!\!
        \left. 
            \frac{\partial \lambda^D_j(\phi)}{\partial \phi} 
        \right|_{\phi^{k}}
    \right.  \\
    & \hspace{13mm} +
    \left.
            \frac{1}{B} \sum_{i=1}^{B}
            \left. 
                \frac{\partial L^{bf}(x_i; \theta)}{\partial \theta} 
            \right|_{\theta^k}
            \!\!\! \cdot \!\!
            \left. 
                \frac{\partial L^{tr}(x^M_j; \theta)}{\partial \theta}  
            \right|_{\theta^{k}}
        \!\!\! \cdot \!\!
        \left. 
            \frac{\partial \lambda^M_j(\phi)}{\partial \phi} 
        \right|_{\phi^{k}}
    \right)\\
    &= - \frac{\eta_\theta}{B} \sum_{j=1}^{B}
    \left(
        G^D(j) \cdot 
        \left. 
            \frac{\partial \lambda^D_j(\phi)}{\partial \phi} 
        \right|_{\phi^{k}} 
        + G^M(j) \cdot 
        \left. 
            \frac{\partial \lambda^M_j(\phi)}{\partial \phi} 
        \right|_{\phi^{k}}
    \right)\\
    &= - \frac{\eta_\theta}{B} \sum_{j=1}^{B}
    G(j) 
    \left.
        \frac{\partial h(\phi)}{\partial \phi}
    \right|_{\phi^{k}},
\end{split} 
\label{eq:derivation_term}
\end{sequation}
where the last term in Eq.~(\ref{eq:derivation_term}) is the derivative of the RRN $h_j(\phi)$ outputs of the $j$-th training sample pair with respect to the network parameters $\phi$, that is
\begin{sequation}
    \frac{\partial h(\phi)}{\partial \phi} = 
    \left[
        \begin{array}{c}
            \left. 
                \frac{\partial \lambda^D_j(\phi)}{\partial \phi} 
            \right |_{\phi^{k}}\\
            \left. 
                \frac{\partial \lambda^M_j(\phi)}{\partial \phi} 
            \right |_{\phi^{k}}
        \end{array}
    \right],
\end{sequation}
and the coefficients $G(j)=\left[ G^D(j) \quad G^M(j) \right]$, where
\begin{sequation}
\begin{split}
    G^D(j) &= 
            \frac{1}{B} \sum_{i=1}^{B}
            \left. 
                \frac{\partial L^{bf}(x_i; \theta)}{\partial \theta} 
            \right |_{\theta^k}
        \cdot
        \left. 
            \frac{\partial L^{tr}(x^D_j; \theta)}{\partial \theta}  
        \right|_{\theta^{k}} \\
        &\triangleq \frac{1}{B} \sum_{i=1}^{B} g^{bf}(x_i) \cdot g^{tr}(x^D_j),\\
    G^M(j) &= 
            \frac{1}{B} \sum_{i=1}^{B}
            \left. 
                \frac{\partial L^{bf}(x_i; \theta)}{\partial \theta}
            \right |_{\theta^k}
        \cdot
        \left. 
            \frac{\partial L^{tr}(x^M_j; \theta)}{\partial \theta}  
        \right|_{\theta^{k}} \\
        &\triangleq \frac{1}{B} \sum_{i=1}^{B} g^{bf}(x_i) \cdot g^{tr}(x^M_j),
\end{split}
\end{sequation}
respectively.
Denote the gradient of the meta loss of the $i$-th sample of $\mathcal{D}^{bf}$ as $g^{bf}(x_i) = \left. \frac{\partial L^{bf}(x_i; \theta)}{\partial \theta} \right |_{\theta^k}$, and the gradient of training loss on the $j$-th sample pair of $\mathcal{D}^{tr}$ as $g^{tr}(x^D_j) = \left. \frac{\partial L^{tr}(x^D_j; \theta)}{\partial \theta} \right|_{\theta^{k}}$ and $g^{tr}(x^M_j) = \left. \frac{\partial L^{tr}(x^M_j; \theta)}{\partial \theta} \right|_{\theta^{k}}$. Obviously, the coefficient $G(j)$ represents the similarity between the gradient of training loss $\left[ g^{tr}(x^M_j) \quad g^{tr}(x^D_j) \right]$ and the average of the gradient of meta loss $g^{bf}(x_i)$. Furthermore, we can reformulate the average gradient of the meta loss by class. The coefficient $G(j)$ can be represented as:
\begin{sequation}
\begin{split}
    G^D(j) &= 
        \frac{1}{B} 
        \sum_{c=1}^{C_t}
        \left(
            \sum_{i=1}^{B_c} g^{bf}(x_i)
        \right)
        g^{tr}(x^D_j),\\
    G^M(j) &= 
        \frac{1}{B} 
        \sum_{c=1}^{C_t}
        \left(
            \sum_{i=1}^{B_c} g^{bf}(x_i)
        \right)
        g^{tr}(x^M_j),
\end{split}
\end{sequation}
where $C_t$ is the number of all seen classes, $B_c$ is the sample number of each class in a batch, and $\sum^{C_t}_{c=1} B_c = B$. This formulation shows that the coefficients $G(j)$ implicitly model the relationship between the knowledge extracted from each training sample of the new task and that from the average meta samples.

Then the first derivative term in Eq.~(\ref{eq:derivation_term}) can be represented as:
\begin{sequation}
\begin{split}
    \triangledown_\phi \mathcal{L}^{bf}(\mathcal{D}^{bf}; \theta(\phi)) 
    = - \frac{\eta_\theta}{B} \sum_{i=1}^{B}
    G(j)
    \cdot
    \left.
        \frac{\partial h(\phi)}{\partial \phi}
    \right |_{\phi^{k}}.
\label{eq:meta_gradient}
\end{split}
\end{sequation}

Therefore, the gradient of the RRN parameters $\phi$ can be calculated by Eq.~(\ref{eq:phi-update}), which can be easily done in PyTorch \cite{pytorch} with the automatic differentiation.

\begin{figure*}[t]
\centering
\subfloat[CIFAR-100 ER-ACE/RER-ACE]{\includegraphics[width=0.46\textwidth]{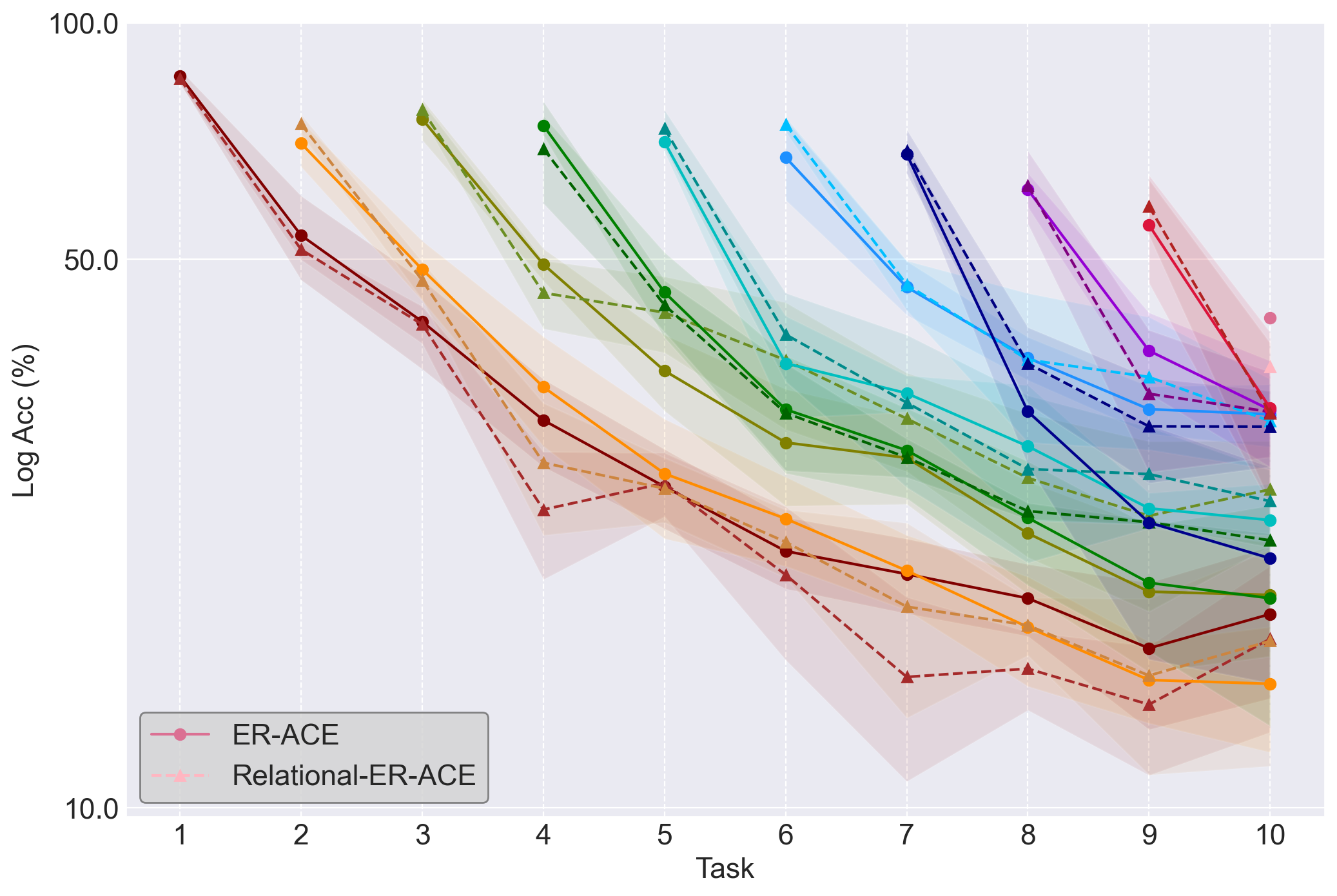}%
\label{subfig:cifar100-er-ace}}
\hfil
\subfloat[Tiny-ImageNet ER-ACE/RER-ACE]{\includegraphics[width=0.46\textwidth]{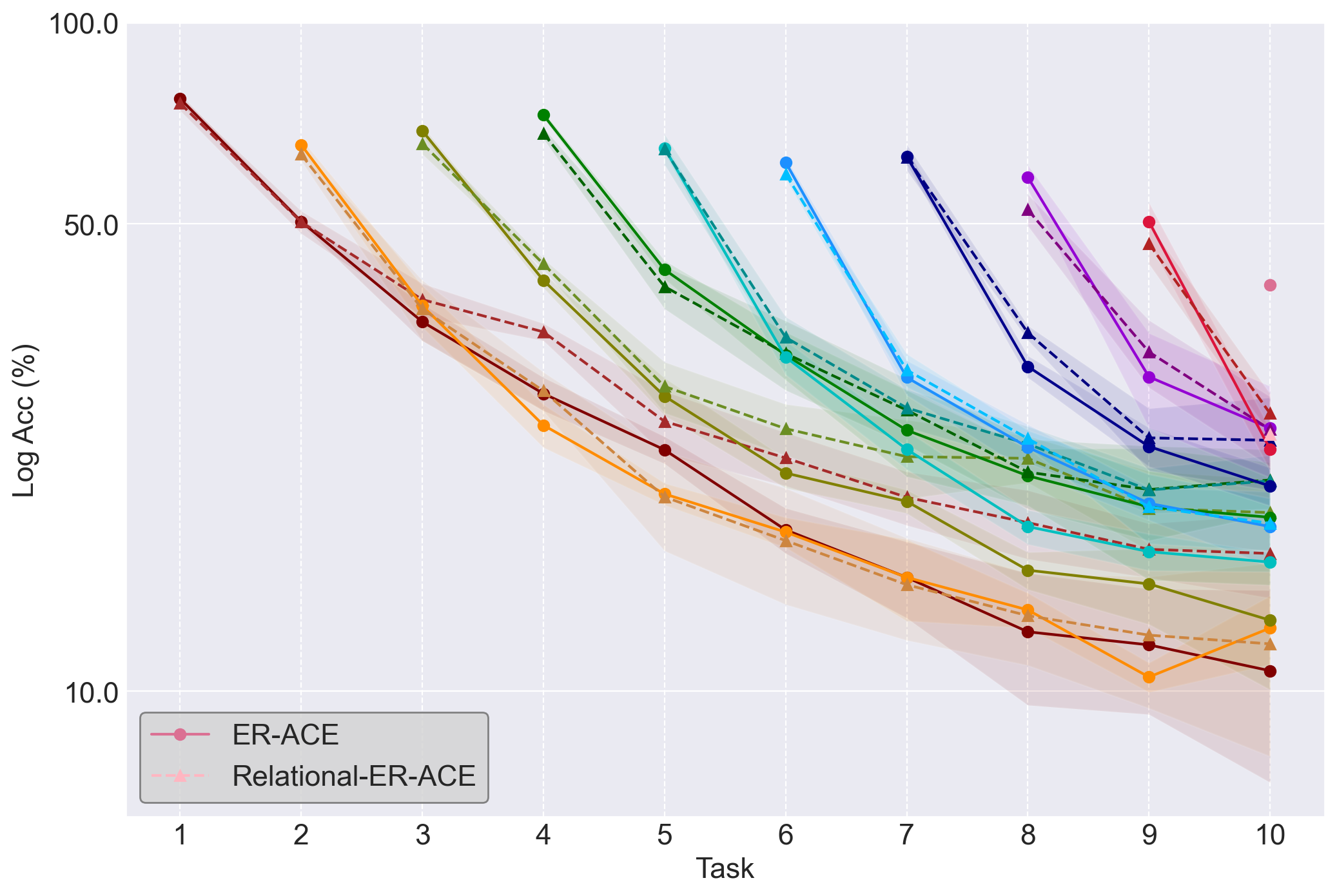}%
\label{subfig:tinyimg-er-ace}}

\subfloat[CIFAR-100 DER++/RDER]{\includegraphics[width=0.46\textwidth]{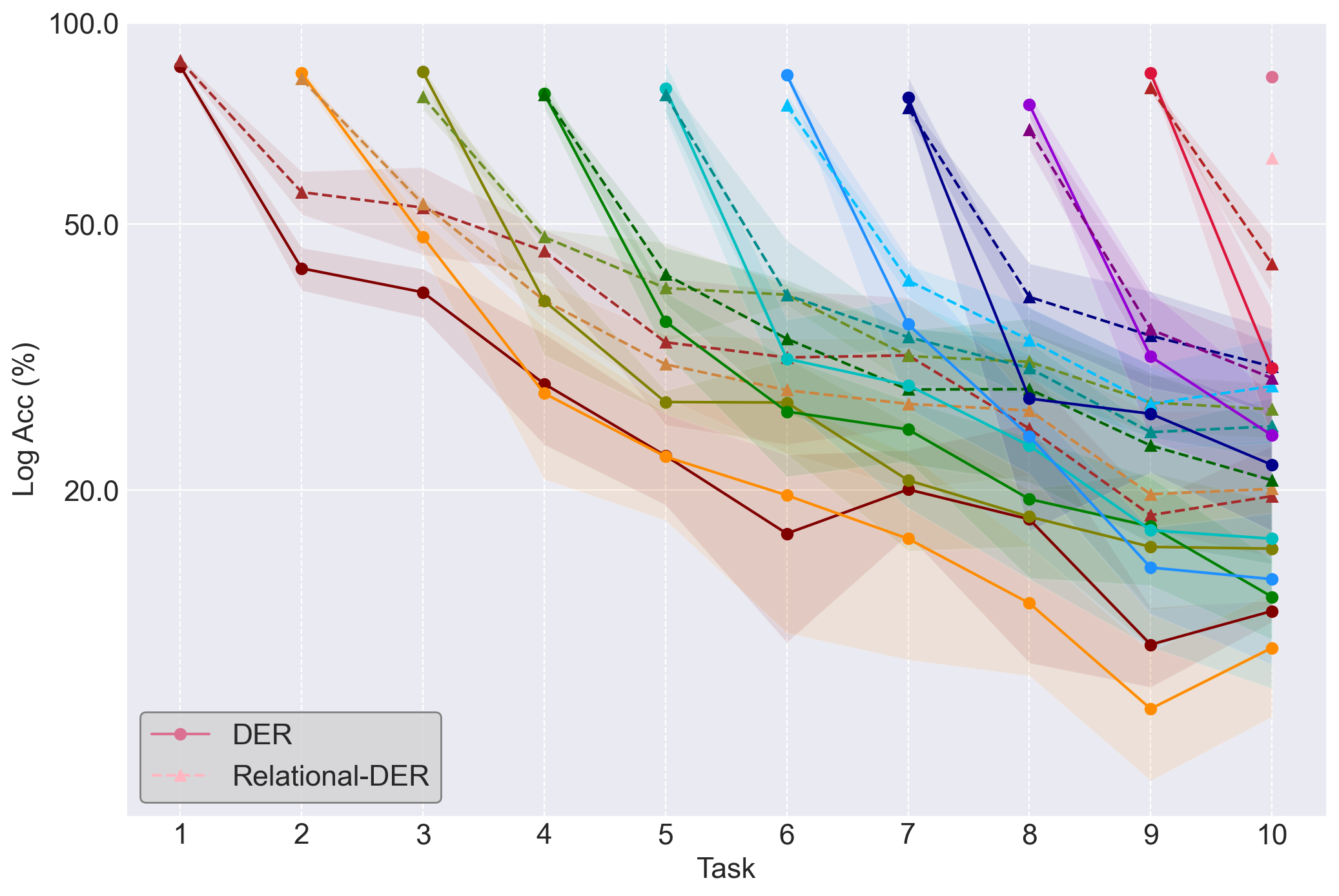}%
\label{subfig:cifar100-der}}
\hfil
\subfloat[Tiny-ImageNet DER++/RDER]{\includegraphics[width=0.46\textwidth]{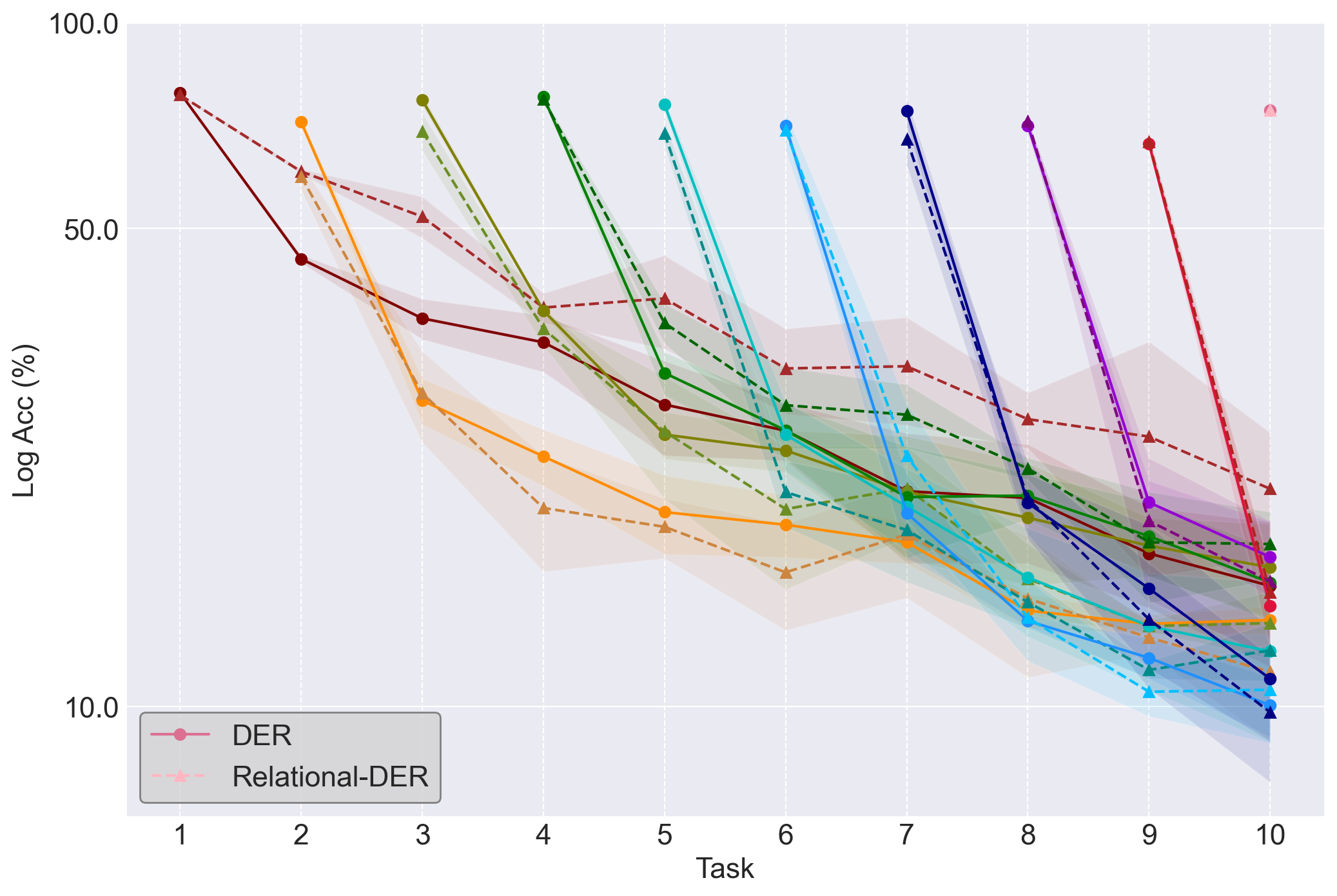}%
\label{subfig:tinyimg-der}}
\caption{Classification accuracy (\%) of each task in the whole training process. The dot-solid lines represent the comparison methods ER, ER-ACE, or DER++ and the triangle-dashed lines represent our methods RER, RER-ACE, or RDER.}
\label{fig:acc_per_task-add}
\end{figure*}

\section{More Experiment Details}
In this section, we first illustrate some experimental details and then present some additional results to demonstrate the effectiveness of our method further.

\subsection{How to Apply the Proposed Method to Other Baselines?}
\label{appendix:apply-other-baselines}
In the main text of the paper, we take ER as an example to illuminate how our proposed approach helps \textit{rehearsal-based} continual learning models deal with the `stability-plasticity' dilemma.
Here we present how to apply our proposed method to other baselines, \emph{i.e.}, ER-ACE~\cite{ER-ACE}, and DER++~\cite{DER}.

\paragraph{Relational-ER-ACE (RER-ACE)}
ER-ACE~\cite{ER-ACE}, which represents `Experience Replay with Asymmetric Cross-Entropy',
combines the losses of the new task samples and the memory buffer samples as
\begin{sequation}
\begin{split}
    \mathcal{L}^{tr}(\theta) = 
    \frac{1}{B} \sum^{B}_{i=1} 
    &\lambda^D {L}_{CE}(x^D_i; \theta, C_{curr}) \\
    + &\lambda^M {L}_{CE}(x^M_i; \theta, C_{curr} \cup C_{curr}),
\label{eq:ER-ACE}
\end{split}
\end{sequation}
where the hyperparameters $\Lambda_D$ and $\Lambda_M$ are preset to 1 in \cite{ER-ACE}. The $\mathcal{L}_{CE}(\mathcal{D}; C)$ is defined as:
\begin{sequation}
    L_{CE}(x; \theta, C) = - \log \frac{z_c(x)}{\sum_{c' \in C} z_{c'}(x)},
\end{sequation}
where the sample $x$ belongs to the $c$-th class and $z_c(x)$ is the $c$-th element of the main classification network output $f(x; \theta)$.

Obviously, it is straightforward to apply our proposed method to ER-ACE, where the RRN still generates the weights $[\Lambda_D, \Lambda_M]$ for each training sample pair.
The outer-loop optimization is the same as Eq.~(\ref{eq:outer-loss}) and the generated sample weights are applied to the inner-loop optimization Eq.~(\ref{eq:ER-ACE}).

\paragraph{Relational-DER (RDER)}
The loss function of DER++ is shown in Eq.~(\ref{eq:DER}), which involves three hyperparameters $[\lambda^D, \lambda^M, \gamma^{M}]$. Intuitively, the outer-loop loss function for the RRN in Eq.~(\ref{eq:outer-loss}) can be reformulated as:
\begin{sequation}
    L^{bf}(x;\theta) = L^{bf}_{CE}(x;\theta) + L^{bf}_{KD}(x;\theta),
\label{eq:outer-loss-der}
\end{sequation}
which is the sum of the CE loss and the distillation loss.

\subsection{Other Hyperparameters}
In the warm-up stage (\emph{i.e.} the epochs before ${Iter}_{warm}$), we use preset weights in the inner loop optimization like previous methods.
Specifically, for RER and RER-ACE, the weight for the CE loss of new and old task samples is preset as 1 and 0.5, respectively. And for RDER, the preset weights are 1, 0.5, and 0.2 for the CE loss of new task samples $\mathcal{L}_{CE}(\mathcal{D}_t)$, the CE loss of memory buffer samples $\mathcal{L}_{CE}(\mathcal{M}_t)$, and the distillation loss of memory buffer samples $\mathcal{L}_{KD}(\mathcal{M}_t)$, respectively.

\subsection{Additional Visualization Results}
\label{appendix:visualization}
Here we show the classification accuracy of each task of ER-ACE/RER-ACE and DER++/RDER on CIFAR-100 in Fig.~\ref{fig:acc_per_task-add}(a) and Fig.~\ref{fig:acc_per_task-add}(c), and on Tiny ImageNet in Fig.~\ref{fig:acc_per_task-add}(b) and Fig.~\ref{fig:acc_per_task-add}(d). Similar to Fig.~\ref{fig:acc_per_task-add}, our method obviously improves the accuracy of the previous tasks. Besides, to balance the `stability' and `plasticity', the RDER achieves an overall higher performance even though may not outperform DER++ on some new tasks in Fig.~\ref{fig:acc_per_task-add}(c). All of these results further demonstrate the effectiveness of our method, which can easily be adapted to multiple \textit{rehearsal-based} methods.

 

\bibliography{bare_jrnl_new_sample4}
\bibliographystyle{IEEEtran}


 




\vfill

\end{document}